\documentclass[10pt,journal,compsoc]{IEEEtran}

\usepackage{times}
\usepackage{epsfig}
\usepackage{graphicx}
\usepackage{amsmath}
\usepackage{amssymb}

\usepackage{color}

\usepackage[dvipsnames]{xcolor}
\colorlet{myorange}{green!10!orange}

\usepackage[inline]{enumitem}

\usepackage{booktabs}
\usepackage{multirow}
\usepackage{pifont}
\usepackage{color}
\usepackage{colortbl}
\definecolor{mygray3}{gray}{.9} 
\definecolor{mygray2}{gray}{.8}
\definecolor{mygray1}{gray}{.7}
\definecolor{myblue}{rgb}{0.61, 0.87, 1.0}
\usepackage{amssymb}
\usepackage{amsfonts,amssymb}

\usepackage{algorithm} 
\usepackage{algorithmic}
\usepackage{soul}

\usepackage{ifpdf}
\ifCLASSOPTIONcompsoc
  \usepackage[nocompress]{cite}
\else
  \usepackage{cite}
\fi

\begin{document}
\title{UTSGAN: Unseen Transition Suss GAN for Transition-Aware Image-to-image Translation}
\author{Yaxin Shi,
        Xiaowei Zhou,
        Ping Liu,       Ivor~W.~Tsang,~\IEEEmembership{Fellow,~IEEE}
        
\IEEEcompsocitemizethanks{\IEEEcompsocthanksitem Y. Shi, P. Liu, I.W. Tsang are with the Center for Frontier AI Research, Agency for Science, Technology, and Research, 138632 Singapore. \protect\\
E-mail: shiyaxin.astar@gmail.com, pino.pingliu@gmail.com, Ivor\_Tsang@cfar.a-star.edu.sg
\IEEEcompsocthanksitem
X. Zhou is with the Australian Artificial Intelligence Institute, University of Technology Sydney, Ultimo, NSW, 2007 Australia.\protect\\
E-mail: zxw.edu@gmail.com,
}
\thanks{Manuscript received Nov 02, 2021}} 

\markboth{Journal of \LaTeX\ Class Files,~Vol.~14, No.~8, August~2015}%
{Shell \MakeLowercase{\textit{et al.}}: Bare Demo of IEEEtran.cls for Computer Society Journals}

\IEEEtitleabstractindextext{%

\begin{abstract}

In the field of Image-to-Image (I2I) translation, ensuring consistency between input images and their translated results is a key requirement for producing high-quality and desirable outputs. 
Previous I2I methods have relied on result consistency, which enforces consistency between the translated results and the ground truth output, to achieve this goal. However, result consistency is limited in its ability to handle complex and unseen attribute changes in translation tasks. 
To address this issue, we introduce a transition-aware approach to I2I translation, where the data translation mapping is explicitly parameterized with a transition variable, allowing for the modelling of unobserved translations triggered by unseen transitions. 
Furthermore, we propose the use of transition consistency, defined on the transition variable, to enable regularization of consistency on unobserved translations, which is omitted in previous works. 
Based on these insights, we present Unseen Transition Suss GAN (UTSGAN), a generative framework that constructs a manifold for the transition with a stochastic transition encoder and coherently regularizes and generalizes result consistency and transition consistency on both training and unobserved translations with tailor-designed constraints. 
Extensive experiments on four different I2I tasks performed on five different datasets demonstrate the efficacy of our proposed UTSGAN in performing consistent translations.

\end{abstract}

}

\maketitle

\IEEEdisplaynontitleabstractindextext
\IEEEpeerreviewmaketitle

\IEEEraisesectionheading{\section{Introduction}\label{sec:introduction}}


\IEEEPARstart{I}{mage}-to-Image (I2I) translation is a fundamental problem in computer vision and graphics, which involves transforming an input image from a source domain $X$ to a target domain $Y$. 
Many computer vision tasks, including image colorization~\cite{DBLP:conf/eccv/ZhangIE16}, face editing~\cite{DBLP:conf/iccv/LiuLWT15}, style transfer~\cite{DBLP:conf/cvpr/GatysEB16}, image inpainting~\cite{DBLP:conf/cvpr/PathakKDDE16} and outpainting~\cite{yang2019very}, can be formulated as I2I translation problems. 
Recently, the use of Generative Adversarial Networks (GANs)~\cite{goodfellow2014generative} has led to progress in I2I translation, with the latest methods producing high-quality translated results that are visually appealing to human eyes.

Conditional I2I translation is an extension of vanilla I2I problems that seeks to achieve controllable translation by providing additional guidance. 
For instance, in face editing, the translated images are expected to exhibit specific facial attributes, such as expressions or hair color, as specified by example images or attribute annotations. 
To ensure the translated results consistently reflect the additional guidance, prior GAN-based I2I methods have employed various strategies to enforce consistency in results. Image reconstruction constraints have been used to explicitly supervise consistency with paired data  \cite{DBLP:conf/cvpr/IsolaZZE17,DBLP:conf/nips/ZhuZPDEWS17}, while cycle-consistency has been leveraged to regularize consistency with unpaired data \cite{DBLP:conf/iccv/ZhuPIE17,DBLP:conf/icml/AlmahairiRSBC18,DBLP:conf/icml/KimCKLK17,DBLP:conf/iccv/YiZTG17}.
Other strategies include explicit attribute annotations of output images and the adoption of attribute-preserving constraints to maintain consistency \cite{DBLP:journals/tip/HeZKSC19,DBLP:conf/cvpr/ChoiCKH0C18,DBLP:conf/iccv/LinWCCL19,DBLP:conf/iclr/TaigmanPW17}. 
We collectively refer to these strategies as ``result consistency'' as they are established based on the translated results.

Previous methods for enforcing result consistency in conditional I2I translation have provided useful strategies to encourage consistent translated results. 
However, we have discovered that these strategies are insufficient to equip a model with good generalization ability, which is necessary to obtain satisfactory translation results for complex attribute changes. Specifically, the constraints in prior result consistency works are only applicable to regularize consistency for observed translations embodied with the training image pairs.
It is impractical to achieve model generalization where new images can be generated from any unseen translations not instantiated/observed during the training phase. 
Consequently, these previous methods tend to obtain unsatisfactory results when facing translations with complex attribute configurations, such as multiple attribute changes.

\begin{figure*}[t]
  \begin{center}
        \includegraphics[width=1\textwidth]{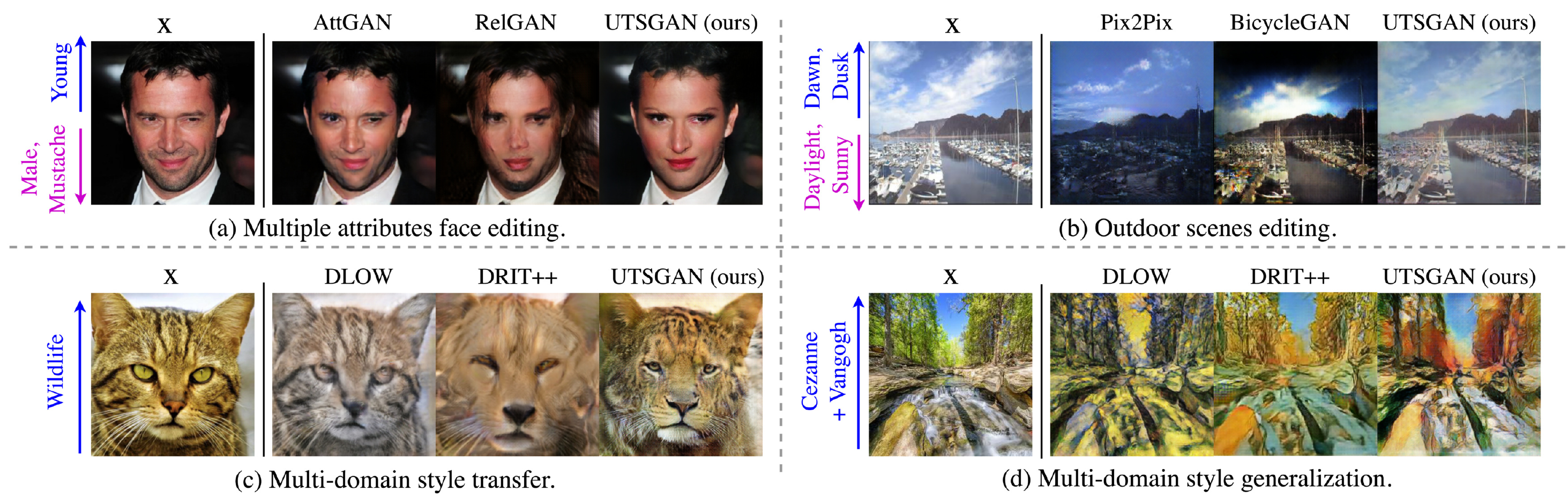}
  \end{center}
  \vspace{-5mm}
\caption{Previous GAN-based I2I translation methods have been found to produce undesirable generation results when performing challenging translations, especially those involving complex configurations such as multiple attribute changes or unobserved multiple attribute/style generalization.}
  \label{fig:motivation_results} 
\vspace{-4mm}
\end{figure*}

In Fig.~\ref{fig:motivation_results}, we present examples to highlight the issue with current result consistency strategies. 
In subfigure(a), AttGAN~\cite{DBLP:journals/tip/HeZKSC19} fails to produce satisfactory results in a multi-attribute face editing task, where enforcing consistency on one attribute leads to undesired artifacts in other attributes. 
Similarly, the result of RelGAN~\cite{DBLP:conf/iccv/LinWCCL19}, which enforces consistency on linearly interpolated attributes, is unrealistic with visible artifacts. 
In subfigure(b), BicycleGAN~\cite{DBLP:conf/nips/ZhuZPDEWS17}, which improves Pix2Pix~\cite{DBLP:conf/cvpr/IsolaZZE17} by using multiple result consistency strategies such as image reconstruction and latent code/attribute regression constraints, still fails to produce desirable results with important image details missing. 
 

One possible solution to address the issue of limited generalization capacity is to collect more data pairs that exemplify a wider variety of attribute changes and regularize result consistency on them~\cite{DBLP:conf/nips/ZhuZPDEWS17}. 
However, this approach is labor-intensive and impractical to exhaustively cover all possible attribute changes.
Alternatively, some works have incorporated intermediate translations defined by a linear coefficient that controls the interpolated proportion between training image pairs, with consistency imposed on these intermediate translations~\cite{DBLP:conf/cvpr/GongLCG19,DBLP:conf/iccv/LinWCCL19}. 
However, due to the limited capacity of this linear parameter, these methods can only enforce consistency on a limited set of interpolated variants of predefined attribute configurations.
Without sufficient numbers of diverse training data pairs and effective consistency regularization on them, previous methods have limited generalization capacity, resulting in poor performance when dealing with unseen data and complex multiple attribute configurations.

We believe that these limitations arise from the fact that the transformation process in prior works tightly coupled with the input-output data pairs ($x\rightarrow{y}$, where {\small$x \in X$} and {\small$y \in Y$} are images from the two domains). 
This formulation restricts models to either rely solely on result consistency regularizers to ensure consistency on observable translations or utilize a confined set of deterministic variants of existing translations for generalization. 
In both cases, the model's generalization ability is highly constrained by the quality and diversity of the training image pairs.

In this work, we introduce the Unseen Transition Suss GAN (UTSGAN) to address the limitations of previous works and facilitate consistent I2I translation. 
UTSGAN explicitly parameterizes each data mapping with a transition variable, resulting in transition-aware I2I translation with three interrelated input variables, namely $x\overset{t}{\mapsto}y$, where $t$ characterizes the desired attribute change from $x$ to $y$. 
This reformulation offers several advantages, such as enabling flexible instantiation of unobserved translations by manipulating the transition variable and allowing for consistency enforcement on the translation process itself, \textit{i.e.}, transition consistency, to ensure the production of consistent translated results. 

{Based on this novel formulation, UTSGAN learns a transition manifold {\small$T$} with a stochastic transition encoder {\small$E(x,y)$}, 
supporting us to flexibly instantiate unobserved translations by triggering arbitrary translations with on-manifold unseen transitions (Fig.~\ref{fig:triplet_matching}).
A holistic set of loss terms is designed to enforce transition consistency on both the observed transitions and our sampled unseen transitions in a coherent manner.
These loss terms complement result consistency, leading to sample-level triplet data consistency for individual translations. 
Finally, triplet distribution matching is performed to generalize our overall regularization of triplet data consistency to distribution-level. 
{In this way, UTSGAN enforces a comprehensive consistency regularization for I2I translation between $X$ and $Y$~\cite{sep-logic-higher-order}, where for {\small$\forall t \in T, \forall x \in X, x \overset{t}{\mapsto}y$} to be true, \textit{i.e.}, {$X \overset{T}{\mapsto}Y$}, must hold (refer to Fig.~2 in Appendix).}



Compared to prior works, UTSGAN possesses outstanding generalization ability to perform consistent I2I translation.
Additionally, explicitly parameterizing the transition data mappings and modelling a generative transition mechanism on them, UTSGAN is general enough to explain existing I2I methods and is capable of handling various I2I translation tasks, with the transition characterized with handcrafted aspects of attribute change.

Experimentally, UTSGAN has shown the capability to achieve satisfactory and consistent translated results in challenging tasks where prior methods failed to produce desirable results.
For example, we achieve the best FID/SSIM scores on multi-attribute face editing (Fig.~\ref{fig:motivation_results}.~(a)) and multi-attribute outdoor scenes editing (Fig.~\ref{fig:motivation_results}.~(b)) tasks. 
We also achieve remarkable results in multi-domain style transfer and generalization tasks, where the output is expected to present one predefined domain style or an unobserved fusion of multiple domain styles. 
As shown in Fig.~\ref{fig:motivation_results}.~(c) and Fig.~\ref{fig:motivation_results}.~(d), compared with other methods, our translated results are outstanding, presenting a harmonious combination of clear image contents and distinguishable desired styles.

To summarize, the contributions of our work are threefold.
\begin{itemize}\setlength{\itemsep}{1mm}
\item We introduce explicit transition-aware I2I translation by parameterizing each data mapping with a transition variable. 
This allows for flexible instantiation of unobserved translations and consistency enforcement on the translation process itself, namely transition consistency, to ensure the production of consistent translated results.

\item We design a transition-aware I2I translation network, \textit{i.e.}, UTSGAN, to coherently regularize and generalize triplet data consistency to distribution-level via triplet distribution matching. 
This enables UTSGAN to achieve outstanding generalization ability and satisfactory and consistent translated results in complex multiple-attribute editing tasks. 
Additionally, parameterizing data mapping makes UTSGAN general enough to explain existing I2I methods and capable of handling various I2I translation tasks with the transition characterized with handcrafted aspects of attribute change.

\item We perform a comprehensive evaluation of UTSGAN to demonstrate its efficacy and generality. 
We evaluate its effectiveness in multi-attribute image editing tasks, including face editing and outdoor scene editing. 
To test its generality, we apply UTSGAN to multi-domain style transfer and generalization and image inpainting tasks. 
Our experimental results indicate that UTSGAN outperforms state-of-the-art methods in all tasks, demonstrating its efficacy and versatility.
\end{itemize}

 \section{Related work}\label{sec:related_work}


{\noindent \textbf{Image-to-Image Translation} involves converting an image from domain {\small$X$} to another domain {\small$Y$}. In practical I2I translation tasks, the translated results are expected to be both visually realistic and semantically consistent, showing desired attribute properties. To achieve this goal, conditional GANs consisting of a generator $G$ and a discriminator $D$, are commonly utilized to model the translation process, \textit{i.e.}, {\small$G: X\mapsto Y$}. 
Although the visual quality of the results is highly improved with powerful data generators, e.g. pre-trained StyleGAN-based generators~{\cite{karras2019style}}, the aspect of consistency in I2I translation has not been systematically studied.}

\begin{figure}[t]
	\centering
 	\includegraphics[width=0.48\textwidth]{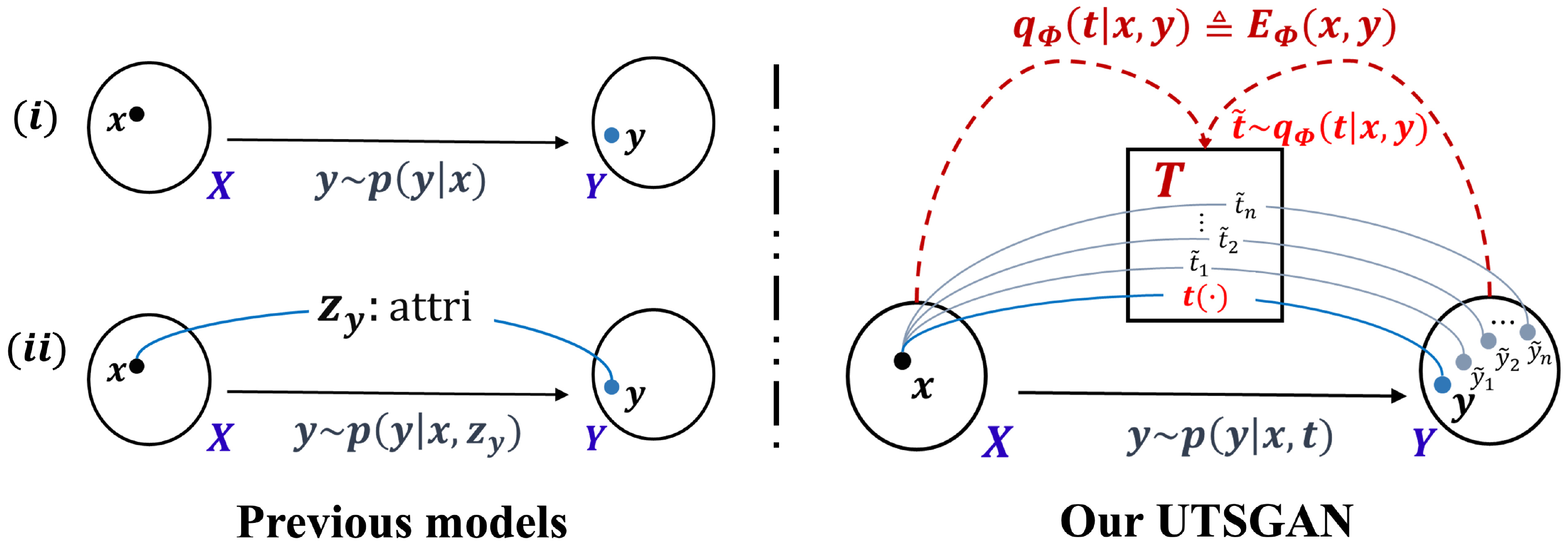}
	\vspace{-3mm}
	\caption{{
Motivation and Illustration for our insight of transition-aware I2I, transition encoding and transition consistency in UTSGAN. 
Previous models focus on maintaining result consistency, with the data mapping implied within  {(\textit{i}) the observed training images or (\textit{ii}) attributes of the target output}, \textit{i.e.}, $z_{y}$.
Failing to ensure consistency for unobserved data mappings, these models lack the generalization ability to achieve high-quality translated results with unseen transitions, \textit{e.g.}, Fig~\ref{fig:motivation_results}. (a) and Fig~\ref{fig:motivation_results}. (d).
Our proposed UTSGAN considers both result consistency and transition consistency, namely, consistency ensured on the data mappings. 
We explicitly parameterize the data mapping with transition {\small$t\triangleq t(x, y)$}, resulting in transition-aware I2I. We model arbitrary unseen transitions through encoding, \textit{i.e.}, {\small$\tilde{t}\sim q(t|x,y)$}. We further generalize the transition consistency defined in observed transitions to unseen transitions with the designed constraints in {Sec.~\ref{sec:UTSGAN_model_design}}. Such a design empowers UTSGAN superior generalization ability to perform consistent I2I translation. 
Moreover, as a general generative I2I translation framework with generative transition, UTSGAN can explain the existing models and is capable to handle various I2I translation tasks, with the transition characterized with handcrafted aspects of attribute change.
   }}\label{fig:utsgan_model}
	\vspace{-5mm}
\end{figure}
{\noindent \textbf{Supervised and Unsupervised I2I Translation.} 
Existing I2I translation methods can be classified into supervised I2I~{\cite{DBLP:conf/cvpr/IsolaZZE17,DBLP:conf/nips/ZhuZPDEWS17,DBLP:conf/icml/AlmahairiRSBC18,DBLP:conf/cvpr/Yu0YSLH18,zheng2019pluralistic,yan2019PENnet}} and unsupervised I2I approaches~{\cite{DBLP:journals/tip/HeZKSC19, DBLP:conf/cvpr/ChoiCKH0C18,DBLP:conf/iccv/LinWCCL19,choi2020stargan,huang2018munit, DBLP:conf/cvpr/GongLCG19,lee2020drit++,luo2021category}}, depending on whether paired training data is available. 
In supervised I2I, exemplar paired input-output images with exact correspondence are used for training, and the goal is to learn a mapping {\small$G: X\mapsto Y$} that is consistent with the given pairs, \textit{i.e.}, 
{\small$G(x)\approx y$}. 
The intuitive result consistency strategy of image reconstruction, \textit{i.e.}, {\small$G(x)=\hat{y}\approx{y}$} is applied in methods like Pix2Pix~\cite{DBLP:conf/cvpr/IsolaZZE17} and BicycleGAN~\cite{DBLP:conf/nips/ZhuZPDEWS17}.

In unsupervised I2I, since paired data is not available, the image reconstruction strategy cannot be adopted any more; other strategies have to be exploited to enforce consistency in the translation.
These strategies include cycle-consistency~\cite{DBLP:conf/iccv/ZhuPIE17,DBLP:conf/icml/AlmahairiRSBC18,DBLP:conf/icml/KimCKLK17,DBLP:conf/iccv/YiZTG17}, attributes preserving constraints~\cite{DBLP:journals/tip/HeZKSC19,DBLP:conf/cvpr/ChoiCKH0C18,DBLP:conf/iccv/LinWCCL19,DBLP:conf/iclr/TaigmanPW17}, and shared latent embedding assumption~\cite{DBLP:conf/nips/LiuBK17,huang2018munit}. 
For example, CycleGAN~\cite{DBLP:conf/iccv/ZhuPIE17},  DiscoGAN~\cite{DBLP:conf/icml/KimCKLK17}, and DualGAN~\cite{DBLP:conf/iccv/YiZTG17} leverage cycle-consistency to enforce consistency between the input and output domains, \textit{i.e.,} {\small$G^{-1}(G(x))\approx x$}, where {\small$G^{-1}(\cdot)$} denotes the inverse mapping of $G$. 
AttGAN~\cite{DBLP:journals/tip/HeZKSC19} and RelGAN~\cite{DBLP:conf/iccv/LinWCCL19} enforce the output image to preserve the desired attributes, \textit{i.e.}, {\small$f(\hat{y})\rightarrow{z_{\hat{y}}}\approx z_{y}$} where {\small$z_{y}$} is the desired attribute of the output image, $f(\cdot)$ is the attribute classifier, and {\small${z_{\hat{y}}}$} is the attribute prediction of $\hat{y}$.
UNIT~\cite{DBLP:conf/nips/LiuBK17} and IcGAN~\cite{DBLP:journals/corr/PerarnauWRA16} rely on a shared latent subspace assumption on the input and output images to enforce result consistency.
{Since these methods rely solely on exemplar data pairs for training, they are limited to utilizing one-to-one observable data mapping to exemplify consistent I2I translation.}
}

\noindent \textbf{Multi-domain and multi-modal I2I Translation.} 
Recently, emerging works have targeted generalized I2I translation with multiple output domains or with multiple diverse outputs, \textit{i.e.}, {one-to-many data mapping}. 
Multi-domain I2I translation aims to achieve consistent translations across multiple output domains. 
{Most of the representative works, \textit{e.g. } ModularGAN~\cite{zhao2018modular}, CircularGAN~\cite{wang2020multi}, and Multi-path consistency GAN~\cite{DBLP:conf/ijcai/LinXWQ019}, still leverage aforementioned result consistency strategies to ensure consistent outputs with cycle-consistency or auxiliary-domain based multi-path consistency regularizers. 
DLOW~\cite{DBLP:conf/cvpr/GongLCG19},  modelling intermediate domains with a continuous coefficient, introduces additional consistency regularizers on the linear interpolation parameter to improve model generalization.}

Multi-modal I2I translation introduces random noise or style codes as a condition for each translation to facilitate one-to-many mapping.
For example, BicycleGAN~\cite{DBLP:conf/nips/ZhuZPDEWS17} and Augmented CycleGAN (AugCGAN)~\cite{DBLP:conf/icml/AlmahairiRSBC18} use random noise as conditions to obtain multiple output images.
MUNIT~\cite{huang2018munit} adopts random style codes to facilitate multiple outputs with diverse styles. 
Representative methods also include SMIT~\cite{romero2019smit}, DRIT~\cite{lee2018diverse}, DRIT++~\cite{lee2020drit++},
and TUNIT~\cite{baek2020rethinking}.
The state-of-the-art methods, such as SDIT~\cite{wang2019sdit}, MMUIT~\cite{liu2021smoothing} and StarGANv2~\cite{choi2020stargan} study multi-domain and multi-modal problem simultaneously.
{These previous multi-modal I2I methods stick to employing result consistency regularizers to enforce consistency on the random factor or instantiated style codes for the sake of consistent I2I translation.}

{\noindent \textbf{High-fidelity I2I Translation.}
    Several specialized works have aimed at enabling GAN-based I2I translation with high-resolution outputs, employing advanced network design and adversarial learning loss for the conditional GAN framework~\cite{zhu2022label,richter2022enhancing,tang2022multi,saharia2022image}. Representative methods in this field include Pix2pixHD~\cite{wang2018high}, SPADE~\cite{park2019semantic}, OASIS~\cite{sushko2020you}, and DeepI2I~\cite{yu2020deepi2i}. These methods prioritize image quality discrimination and semantic-label injection using improved network structure design and more advanced regularization techniques, such as the LabelMix regularization in OASIS. However, these strategies still fall under the result consistency category.}

\noindent \textbf{Limitations {in terms of consistent I2I.}}
To encourage consistent translated results, prior works, as outlined above, mainly resort to the result consistency strategies, \textit{i.e.},
leveraging data relations (input-output correspondence) or data attributes to form consistency constraints on the translated results. 
However, since these constraints are straightforwardly imposed on the translated results, they heavily rely on translations embodied within observed data pairs to regularize while being unable to enforce consistency for the massive translations unobserved during training. In this way, previous I2I works suffer from certain limitations: 

\begin{itemize}\setlength{\itemsep}{0.3mm}
\item Those works focusing on one-to-one mappings, such as Pix2Pix~\cite{DBLP:conf/cvpr/IsolaZZE17}, CycleGAN~\cite{DBLP:conf/iccv/ZhuPIE17}, AttGAN~\cite{DBLP:journals/tip/HeZKSC19}, can only enforce consistency on translations predefined by the training data or given attribute conditions.
These translations can merely exemplify fixed attribute changes or predefined linear variants of them, as in RelGAN~\cite{DBLP:conf/iccv/LinWCCL19}.
These methods are hardly capable of ensuring consistency on new translations configured with unseen data or unseen attribute changes.

\item The works studying one-to-many mappings, including BicycleGAN~\cite{DBLP:conf/nips/ZhuZPDEWS17}, AugCGAN~\cite{DBLP:conf/icml/AlmahairiRSBC18}, DLOW~\cite{DBLP:conf/cvpr/GongLCG19}, StarGAN~\cite{DBLP:conf/cvpr/ChoiCKH0C18}, StarGANv2~\cite{choi2020stargan}, and {DRIT++~\cite{lee2020drit++}}, go beyond predefined translations/conditions by incorporating various output attributes, such as random factors or latent codes, to train their models. Seeking variation on model output, they still leverage result consistency regularizers to enforce consistency on those new translations.
However, since the variation reflected in the ultimate results is of low granularity,
their exemplified translations can only contribute limited translation variations, failing to cover the massive unseen attribute changes. This makes it difficult for them to achieve sufficient model generalization ability for consistent I2I translation.
\end{itemize}

In this paper, we propose a method to overcome the limitations of previous works, which had deficiencies in model generalization ability and obtained inconsistent or unsatisfactory translated results when handling translations with complex attribute configurations.
Our approach involves explicitly parameterizing the transition mapping with the change of condition, thus removing the restriction posed by predefined conditions. 
We argue that by enforcing and generalizing consistency defined on the transition mapping itself, we can ensure that the learned translation function generally preserves the desired attribute or condition change indicated by the arbitrarily designed transition mapping characterization.

It should be noted that our work primarily focuses on exploring regularizations to facilitate consistent I2I translation. 
Therefore, high-fidelity I2I works that achieve consistent translation through strong semantic-label injection network structures are not within the scope of our main objective, and we left them as future explorations.

\section{Methodology}

In this section, we first elaborate on our reformulation of transition-aware I2I translation and transition consistency in Sec.~\ref{sec:transition-aware I2I}.
Then, we present our designed \textbf{U}nseen \textbf{T}ransition \textbf{S}uss GAN (UTSGAN) for consistent I2I translation, in Sec.~\ref{sec:UTSGAN}.

\subsection{Transition-aware I2I translation}\label{sec:transition-aware I2I}
To model the I2I translation process in a more flexible manner, we introduce a transition variable $t$ to explicitly parameterize the mapping operation embedded within each input-output data pair, denoted as $t\triangleq t(x,y)$. 
By including the transition variable, we can reformulate the I2I translation problem as a transition-aware image generation problem with three interrelated inputs represented as $x \overset{t}{\mapsto} y$. 
This formulation provides more flexibility in modelling the translation process by allowing us to go beyond the observable mappings predefined by the training output images or given attribute conditions. 
By manipulating $t$ to $\tilde{t}$, we can readily model unobserved translations by learning from the inputs $x$ and $\tilde{y}$, where $x \overset{\tilde{t}}{\mapsto}\tilde{y}$. 



Based on the above formulation, we have introduced the concept of ``transition consistency'', which refers to consistency in terms of the data mapping process. 
Specifically, if two translated results share the same attribute(s), their corresponding mapping operations (transitions) should be the same or similar, ensuring consistency.
{Approaching consistency on the relative change of paired images/conditions through translation, instead of on each individual translated result, we ease the requirement on the diversity of predefined conditions for model generalization.}

For the convenience of further discussions, we provide the formal definition of result consistency and transition consistency {under the transition-aware/conditioned I2I formulation} as follows:
\begin{itemize}

\item[{\small {\small (1)}}] \textit{Result consistency} is defined as consistency defined on the output images. 
In previous I2I works, it is ensured by self-reconstruction, which requires that {\small$G(x,0)={\hat{x}}\approx x$}; cycle consistency, which requires that {\small$G(G(x,t),-t) \approx x$}. In a supervised setting, it can be ensured by {\small$G(x,t)={\hat{y}}\approx y$}.


\item[{\small (2)}] 

\textit{Transition consistency} is defined as the consistency of transition mappings. 
Let $t(\cdot)$ be an arbitrary transition prediction function for input-output image pairs. 
Let {\small$\hat{t}_{y}$} and {\small$\hat{t}_{\hat{y}}$} be the transition prediction results of data pairs {\small${x,y}$} and {\small${x,\hat{y}}$}, respectively, \textit{i.e.}, {\small$t(x,y)\rightarrow \hat{t}{y}, t(x,\hat{y})\rightarrow \hat{t}_{\hat{y}}$}. 
With transition consistency, we require that if $y \approx \hat{y}$, then $\hat{t}_{y}\approx{\hat{t}_{\hat{y}}}$.
\end{itemize}

We summarize the key notations in Table.~\ref{tab:notations}.

Incorporating transition consistency into I2I translation problems has several advantages. 
Firstly, it is applicable to both supervised and unsupervised problem settings since it does not rely on paired data to regularize. 
Secondly, it promotes model generalization by enforcing consistency on unobserved translations that arise from unseen transitions. 
Thirdly, it operates at a higher granularity than result consistency 
{since it is defined based on the data mapping process, rather than on the ultimate translated results.} 
Therefore, it enables more precise consistency regularization for individual translations.
	\begin{table*}[t]
		\begin{center}
			\renewcommand{\arraystretch}{1.2}
     \caption{Mathematical notations in our work. (Presented in order)}~\label{tab:notations}
     \vspace{-4mm}
			\setlength{\tabcolsep}{1mm}{
				\scalebox{1}{
					\begin{tabular}{cl}
						\toprule 
          \textbf{Notation} & \textbf{Description}\\ \hline
         $\{x,y\}$, where $x\in X$, $y \in Y$    & arbitrary training image pairs (either paired or unpaired) from the source domain and target domain in I2I translation. 
        \\  {\small$x\; \mapsto \; y$}          & traditional I2I translation with arbitrary observed translation embodied within training data pairs.

        \\ {\small$x\; \overset{ t\triangleq t(x,y)}{\mapsto} \; y$} & transition-aware I2I translation where the observed translations are additionally modelled with an explicit transition variable.
        \\{\small$G(x,0)={\hat{x}}$} & self-reconstruction of ${x}$. 
        \\{\small$G(x,t)={\hat{y}}$}  & transition conditioned generation of $\hat{y}$, based on the ground truth transition $t$.
        \\{\small $\tilde{t}$} & {unseen transitions}, variants of the observed transition $t$.
        \\{\small$G(x,\tilde{t})={\tilde{y}}$}  & transition conditioned generation of $\tilde{y}$, based on the established transition variants $\tilde{t}$.
         \\ {\small$x\; \overset{\tilde{t}}{\mapsto} \; \tilde{y}$} & {unobserved translations triggered with unseen transitions.} 
        \\ {\small $t(x,y_{*})\rightarrow \hat{t}_{y_{*}}$} & transition prediction result of arbitrary image pair $\{x, y_{*}\}$.
          \\ \hline
		\bottomrule
			\end{tabular}}}
		\end{center}
		     \vspace{-4mm}
	\end{table*}

\subsection{{U}nseen \textbf{T}ransition \textbf{S}uss GAN}~\label{sec:UTSGAN}
In this section, we introduce the Unseen Transition Suss GAN (UTSGAN) model, which aims to improve model generalization by emphasizing the regularization and generalization of transition consistency on unobserved translations. 
We provide a detailed description of our model design in Sect.~\ref{sec:UTSGAN_model_design}, the objective function in Sect.~\ref{sec:UTSGAN_objective}, and the training procedure in Sect.~\ref{sec:UTSGAN_model_training}.

\subsubsection{Model design}\label{sec:UTSGAN_model_design}

Our UTSGAN model consists of five modules, as shown in Fig.~\ref{fig:UTSGAN_network_simple}: 
the transition encoding module $E$, the transition quality discriminator $D_T$, the I2I translation module $G$,
the image reality/quality discriminator $D_{\text{Real}}$ and the triplet data joint distribution matching discriminator $D_{\text{Match}}$.
{The overall mechanism of UTSGAN involves the interaction between two generative processes jointly supported by these five modules, including the transition-conditional I2I translation process and our proposed transition encoding process.
The primary objective of this mechanism is to achieve high-quality image translation while ensuring consistency with respect to transition mappings. }

\noindent {\textbf{Transition encoding.}} 
{To instantiate diverse unobservable translations, UTSGAN, depicted in Fig.~\ref{fig:utsgan_model}, incorporates a transition encoding module that not only performs transition prediction but also stochastically generates unseen transitions.} 
Specifically, we model the data mapping between {\small$x\mapsto y$} with a {stochastic transition encoder}~\cite{DBLP:journals/corr/KingmaW13}, denoted as {\small$E(x,y)$}. Such a transition encoder can predict transitions for individual data pairs, \textit{i.e.},
\begin{equation}
    E(x,y)\rightarrow \hat{t}_{y}. \qquad 
\end{equation}
It also contributes to learning a distribution of transition $t$ with a $q$ function, \textit{i.e.}, {{\small $q_{{\phi}}(t|x,y)\triangleq E_{{\phi}}(x,y)$}}~\cite{DBLP:journals/corr/KingmaW13}, with our introduced discriminator {\small$D_{T}$} for ensuring the quality of the transitions. 
As depicted in Fig.~\ref{fig:triplet_matching}, the learned distribution characterizes a latent manifold {\small$T$} for transition, where the observed transitions $t$ and all reasonable unseen transitions {\small$\hat{t}_{y}$} lie on. By formulation, it is: 
\begin{equation}
\begin{aligned}
q_{\phi}(t|x,y)&\approx{p(t)}\approx{\mathcal{N}(0,{I})}, \\ 
\tilde{t}&\sim q_{\phi}(t|x,y),
\label{eq:learnable_t}
\end{aligned}
\end{equation}
where {\small$\mathcal{N}(0,{I})$} is the Gaussian prior 
distribution 
for $t$.
{\small$\tilde{t}$} are the unseen transitions randomly sampled from the learned distribution. Note that, the transition prediction results of instanced image pairs should also lie on our  learned transition manifold {\small$T$}.

{Our transition encoding module facilitates flexible manipulation of unseen transitions by allowing for both reasonable and diverse operations. 
As depicted in Fig.~\ref{fig:triplet_matching}, by sampling on-manifold transitions from the encoding, our instantiated unseen transitions have two properties. 
First, they can embody arbitrary reasonable attribute configurations since on-manifold transitions can capture the relation of different data mappings under multiple attribute expectations. 
Second, by modelling transition variants $\tilde{t}$ with uncertainty, our unseen transitions are not confined to deterministic variants predefined by existing transitions, such as linear interpolation. 
Therefore, our modelled unseen transitions are much more varied and have a higher chance of covering complex attribute configurations that were unobserved during training.}
\begin{figure}[t]
  \begin{center}
        \includegraphics[width=0.43\textwidth]{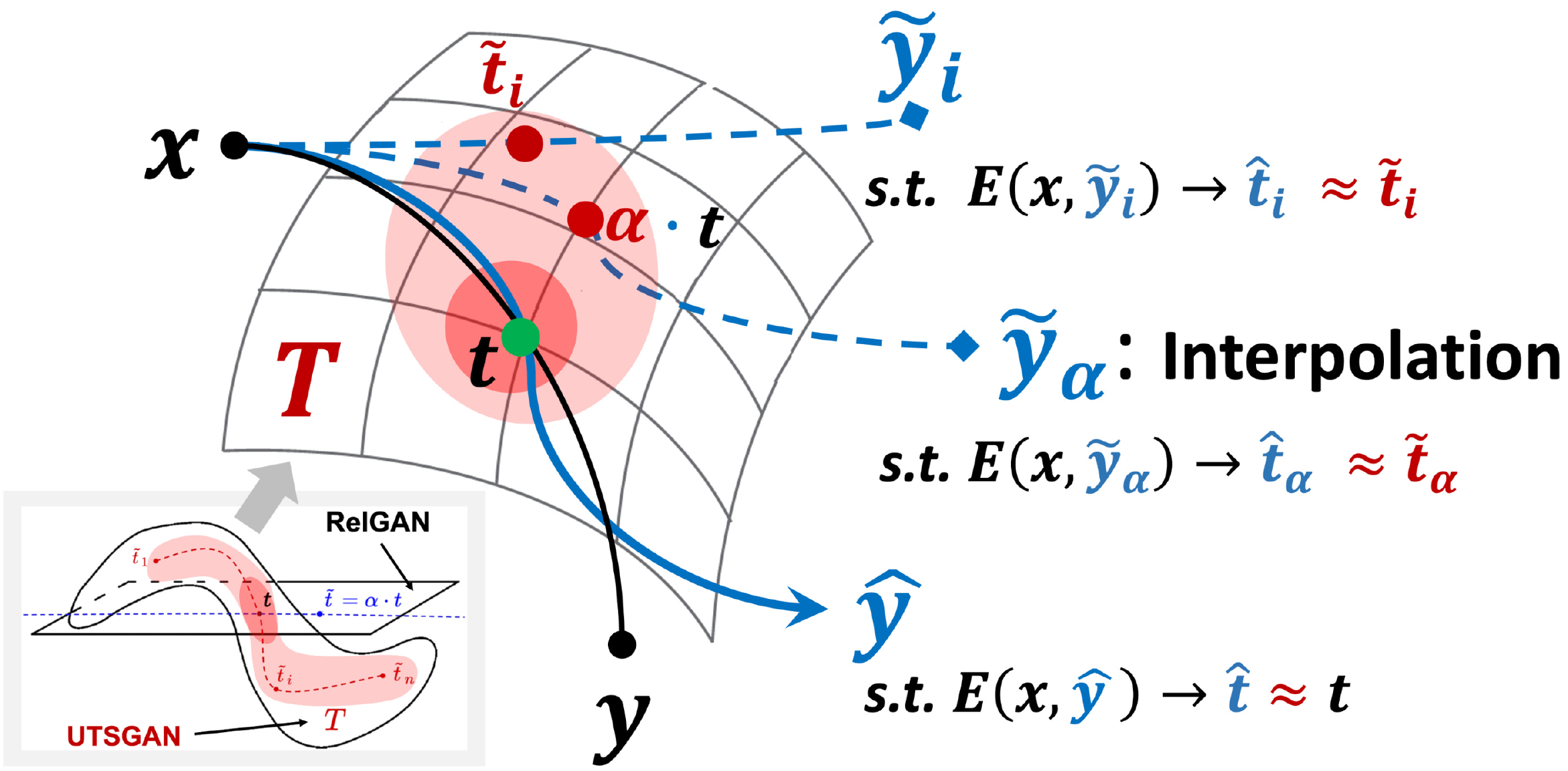} 
  \end{center}
\vspace{-4mm}
    \caption{
    Transition consistency regularization in UTSGAN.
    We learn a distribution for $t$ with transition encoding, \textit{i.e.}, Eq.~\eqref{eq:learnable_t}, leading to a latent manifold denoted as $T$.
    We randomly sample {\small$\tilde{t} \in T$} for each $t$, and regularize consistency between {\small$\tilde{t}$} and its corresponding generated result {\small$\tilde{y}$}, with Eq.~\eqref{eq:sample_consistent_t}. We then generalize such transition consistency, together with the result consistency, to distribution-level with Eq.~\eqref{eq:consistent_t}. Our design enables explicit regularization of consistency, more specifically transition consistency, on unobserved translations. {Foremost, it facilitates flexible and reasonable manipulation of data mappings with diverse operations {under uncertainty}. Interpolation used in previous works is a case of our design with deterministic linear operation on predefined mappings.}
    }\label{fig:triplet_matching}
    \vspace{-5mm}
 \end{figure}
  \begin{figure*}[t]
	\centering
 	\includegraphics[width=0.98\textwidth]{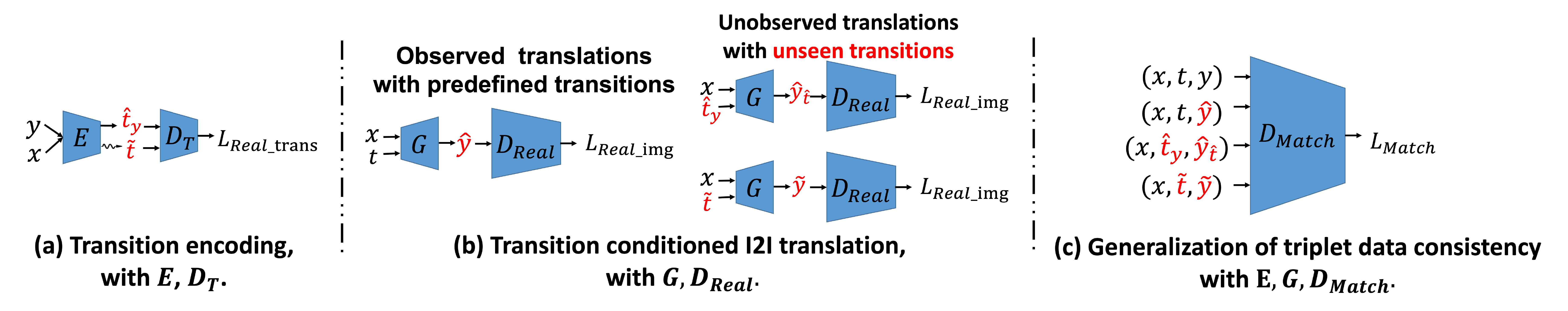}
	\vspace{-4mm}
	\caption{{The illustration for the overall model design of our \textbf{U}nseen \textbf{T}ransition \textbf{S}uss GAN.} The model comprises five modules:  $E, D_{T}, G, D_{\text{Real}}$ and $D_{\text{Match}}$. The variables in red are elements of unobserved translations, \textit{i.e.}, unseen transitions or their corresponding translated results. }\label{fig:UTSGAN_network_simple}
	\vspace{-4mm}
\end{figure*}

\noindent \textbf{{Transition-aware I2I translation.}} 
UTSGAN models the data translation process through a collaborative interaction between the generator $G$ and the transition encoder $E$, as depicted in Fig.~\ref{fig:utsgan_model}. By incorporating the {aforementioned} transition encoding procedure, UTSGAN can model both observed translations, which are predefined by the training data, and unobserved translations triggered by unseen transitions.

\noindent \textbf{-- Observed translations:} Based on the essence of transition as data mapping, for each observed data triplet $(x,t,y)$, we could handily have three seen translations predefined for training.
\begin{equation}
    {\hat{y}}=G(x,{t}) \quad {\hat{x}_{\text{self}}}=G(x,0) \quad \hat{x}_{\text{cycle}}=G(G(x,t),-t). 
\end{equation}

Result consistency defined on these observed translations includes:
\begin{equation}
\begin{aligned}
    \hat{y}\approx y, \quad {\hat{x}_{\text{self}}}&\approx x, \quad {\hat{x}_{\text{cycle}}}\approx x;
\end{aligned}\label{eq:result_consistency}
\end{equation}
where $\hat{y}\approx y$ is enforced only if trained with paired data. 

Transition consistency defined on these translations is:
\begin{equation}
\begin{aligned}
 \hat{t}_{y}&\approx{\hat{t}_{\hat{y}}}\approx t
\end{aligned}
\label{eq:transition_consistency}
\end{equation}
\noindent where {\small$E(x,\hat{y})\rightarrow \hat{t}_{\hat{y}}$}, {\small$E(x,y)\rightarrow \hat{t}_{y}$}, and $t$ is the ground truth. 

\noindent \textbf{-- {Unobserved translations}:} 
Assuming that $x$ is the input image, and $\hat{y}$ and $\tilde{y}$ are the two outputs corresponding to the unseen transitions $\hat{t}_{y}$ and $\tilde{t}$ respectively, we can obtain the unobserved translations as follows:
\begin{equation}
    \hat{y}_{\hat{t}}= G(x,\hat{t}_{y}) \qquad \quad \tilde{y}=G(x,\tilde{t}).
\end{equation}

In this way, we obtain two sets of triplet data that represent example unobserved translations, \textit{i.e.}, {\small$\{x,\hat{t}_{y},\hat{y}_{\hat{t}}\}$} and {\small$\{x,\tilde{t}_{y},\tilde{y}_{\tilde{t}}\}$}.

We can then use these triplet data to regularize transition consistency, thereby establishing a coherent interaction between $E$ and $G$, which is defined as follows:
\begin{equation}
        \begin{aligned}
        E(x,G(x, E(x,y)))\rightarrow\hat{t}_{\hat{y}}, \;\;\;\; &\hat{t}_{\hat{y}} \approx t;\\
        E(x,\tilde{y})\rightarrow\hat{t}_{\tilde{y}},\qquad  \qquad  & \hat{t}_{\tilde{y}}\approx{\tilde{t}};
        \end{aligned}\label{eq:sample_consistent_t}
        \end{equation}        
where {{\small$\hat{t}_{\hat{y}}$}}, {{\small$\hat{t}_{\tilde{y}}$} are the transition prediction results of {\small$(x, \hat{y})$}, {\small$(x,\tilde{y})$}}, respectively.
The quality of the generated images, including $\hat{y}_{t}$,  $\hat{y}_{\hat{t}}$ and $\tilde{y}_{\tilde{t}}$, is evaluated by the discriminator $D_{\text{Real}}$ to ensure that they are realistic and of high quality.


\vspace{1mm}
\noindent \textbf{Generalize triplet data consistency to distribution-level.}
With the explicitly regularized sample-level consistency defined in Eq.~\eqref{eq:result_consistency} and Eq.~\eqref{eq:sample_consistent_t},
we further enforce joint distribution matching on the triplet data, to coherently generalize the established overall consistency on triplet data, \textit{i.e.}, both result consistency and transition consistency, to
distribution-level~\cite{DBLP:conf/cvpr/ZhengZZL0020}, with $D_{\text{Match}}$.
        \begin{equation}
        \begin{aligned}
         {p_{\phi}(x, t, \hat{y})}\approx {p_{\phi}(x, \hat{t}_{y}, \hat{y}_{\hat{t}})}\approx  {p_{\phi}(x, \tilde{t}, \tilde{y})}\approx{p(x,t,y)}\label{eq:consistent_t}
        \end{aligned}
        \end{equation} 
\subsubsection{Objective}\label{sec:UTSGAN_objective}
The objective of UTSGAN consists of the following loss terms. 
\vspace{1mm}\\
\noindent\textbf{\textit{(1).} Loss terms for the {GAN-based image translation}.}  Our loss terms for the translation process of UTSGAN are:

\vspace{1mm}
\noindent\textbf{-- Image reconstruction losses for result consistency:} Based on Eq.~\eqref{eq:result_consistency},   
we have three loss terms for \textit{result consistency}, {which include
self-reconstruction, cycle-reconstruction/consistency and result-reconstruction of the observed training image pairs. }\\
\vspace{1mm}
\noindent The {self-reconstruction loss} is:
\begin{equation}
    \mathop{\min}\limits_{G}{\mathcal{L}_{\text{Recons\_img\_self}}}  = \mathbb{E}_{x\sim{p(x)}}[\parallel G(x,0)-x\parallel_{1}];\label{eq:self_reconstruction}
\end{equation}
\noindent The {cycle-reconstruction loss is}: 
 \begin{equation}
    \mathop{\min}\limits_{G}{\mathcal{L}_{\text{Recons\_img\_cyc}}} = \mathbb{E}_{(x,t)\sim{p(x,t)}}[\parallel G(G(x,t),-t)-x\parallel_{1}]; \label{eq:cycle_reconstruction}
\end{equation}       
\noindent The {result-reconstruction loss} for 
supervised I2I cases is: 
\begin{equation}
    \mathop{\min}\limits_{G}{\mathcal{L}_{\text{Recons\_img\_out}}} = \mathbb{E}_{(x,t,y)\sim{p(x,t,y)}}[\parallel G(x,t)-y\parallel_{1}]. \label{eq:out_reconstruction}
\end{equation}
\vspace{1mm}
\noindent\textbf{-- The adversarial loss for realistic image generation:}
\begin{equation}
\begin{aligned}
\min\limits_{G}\max\limits_{D_{\text{Real}}}\ &\mathcal{L}_{\text{Real\_img}} = \mathbb{E}_{y\sim{p(y)}}[\log D_{\text{Real}}(y)] \\
+&\mathbb{E}_{(x,t)\sim{p(x,t)}}[\log (1- D_{\text{Real}}(G(x,t))] \label{eq:img_generation},
\end{aligned}
\end{equation}
where {the generator $G$} aims to generate realistic images to fool discriminator $D_{\text{Real}}$, and {the discriminator $D_{\text{Real}}$} aims to distinguish the generated images from real ones.

\vspace{1mm}
\noindent\textbf{\textit{(2).} Loss terms for transition consistency.} Our proposed loss terms for \textit{transition consistency} include:
\vspace{1mm}

\noindent\textbf{-- Transition {prediction} loss on observed transitions:} Based on Eq.~\eqref{eq:transition_consistency}, the loss term for transition consistency on predefined/observed transitions can be defined as:
\begin{equation}
\begin{aligned}
\min\limits_{{G},E}{\mathcal{L}_{\text{Recons\_trans}}}  &= \mathbb{E}_{(x,t,y)~\sim{p(x,t,y)}}[\parallel E(x,y)-t\parallel_{1}] \\
\qquad &+\mathbb{E}_{(x,t,y)~\sim{p(x,t,y)}}[\parallel E(x,G(x,t))-t\parallel_{1}]\\
\qquad &+\mathbb{E}_{x~\sim{p(x)}}[\parallel E(x,x)-0\parallel_{1}]\label{eq:transition_recons}
\end{aligned}
\end{equation}

\vspace{1mm}
\noindent\textbf{-- The adversarial loss for unseen transition generation:}
We adopt {discriminator $D_{T}$}, to learn {a distribution for $t$} with our encoder, \textit{i.e.}, Eq.\eqref{eq:learnable_t}, which thus facilitates the generation of new transitions, \textit{i.e.}, \mbox{{\small$\tilde{t}\sim{q(t|x,y)}\triangleq E(x,y)$}}. 
\begin{equation}
\begin{aligned}
\min\limits_{E}\max\limits_{D_{{T}}}\mathcal{L}_{\text{Real\_newtrans}} &= \mathbb{E}_{t\sim {p(t|x,y)}}[\log D_{T}(t)]\\
\qquad & + \mathbb{E}_{t'\sim{\mathcal{N}(0,I)}}[\log D_{T}(t')]\\
\qquad &+ \mathbb{E}_{\tilde{t}\sim{q(t|x,y)}}[\log (1- D_{T}(\tilde{t})]\label{eq:transition_generation},
\end{aligned}
\end{equation}
where $t$ is the seen (realistic) transitions, $t'$, $\tilde{t}$ are transitions sampled from the prior and our learned distribution~\cite{DBLP:conf/aaai/ZhaoSE19}. 
\vspace{1mm}

\noindent\textbf{-- Transition {prediction} loss on unseen transitions:} Based on Eq.~\eqref{eq:sample_consistent_t},
this loss term is defined as:
\begin{equation}
\begin{aligned}
\mathop{\min}\limits_{{G},E}{\mathcal{L}_{\text{Recons\_newtrans}}} & = \mathbb{E}_{\substack{x~\sim{p(x)}\\t'\sim{\mathcal{N}(0,I)}}}{[\parallel E(x,G(x,t'))-t'\parallel_{1}]} \\
\qquad &+\mathbb{E}_{\substack{x~\sim{p(x)}\\\tilde{t}~\sim{q(t|x,y)}}}{[\parallel E(x,G(x,\tilde{t}))-\tilde{t}\parallel_{1}]} \label{eq:new_transition_recons}
\end{aligned}
\end{equation}
\vspace{1mm}
\noindent\textbf{\textit{(3).} Loss terms for generalization of triplet consistency.} 
We first instantiate new translations with those stochastically sampled unseen transitions, \textit{i.e.}, \mbox{{\small$\tilde{y}=G(x,\tilde{t})$}} and \mbox{{\small${y}'=G(x,{t'})$}}. 
\vspace{1mm}


\noindent\textbf{-- The adversarial loss for realistic image generation with unseen transitions:}
\begin{equation}
\begin{aligned}
\min\limits_{G}\max\limits_{D_{\text{Real}}}\ &\mathcal{L}_{\text{Real\_newimg}} = \mathbb{E}_{y\sim{p(y)}}[\log D_{\text{Real}}(y)] \\
+&\mathbb{E}_{x\sim{p(x)},\tilde{t}\sim{q(t|x,y)}}[\log (1- D_{\text{Real}}(G(x,\tilde{t}))] \\
+&\mathbb{E}_{x\sim{p(x)},{t'}\sim{N(0,1)}}[\log (1- D_{\text{Real}}(G(x,{t'}))]  \label{eq:newimg_generation}
\end{aligned}
\end{equation}
\vspace{1mm}
\noindent\textbf{-- The adversarial loss for generalizing {triplet data} consistency to distribution-level:} We adopt a discriminator $D_{\text{Match}}$ that takes triplet inputs, \textit{i.e.}, $(x,t,y)$ to achieve our joint distribution matching design mentioned in Eq.~\eqref{eq:consistent_t}. The loss term is defined as follows:
\begin{equation}
\begin{aligned}
\min\limits_{G}\max\limits_{D_{\text{Match}}}\mathcal{L}_{\text{Match}} = &\;\;\mathbb{E}_{(x,t,y)\sim{p(x,t,y)}}[\log D_{\text{Match}}(x,t,y)] \\
+\mathbb{E}_{(x,t)\sim{p(x,t)}}&[\log (1- D_{\text{Match}}(x, t, G(x,t))] \\
+\mathbb{E}_{x\sim{p(x)},\tilde{t'}\sim{N(0,1)}}&[\log (1- D_{\text{Match}}(x, \tilde{t'}, G(x,\tilde{t'}))]\\
+\mathbb{E}_{x\sim{p(x)},\tilde{t}\sim{q(t|x,y)}}&[\log (1- D_{\text{Match}}(G(x,\tilde{t},G(x,\tilde{t})))] \label{eq:triplet_generalization}
\end{aligned}
\vspace{2mm}
\end{equation}

Based on the proposed method of incorporating random wrong triplets, as described in~\cite{DBLP:conf/iccv/LinWCCL19}, we introduce $(x, t_{\times}, y)$ and $(x, t, y_{\times})$ as random wrong triplets for each real paired triplet $(x, t, y)$ to 
upgrade the triplet data matching process. This results in an updated adversarial distribution loss defined as {\small$\mathcal{L}^{D}_{\text{Match}}$} and {\small$\mathcal{L}^{G}_{\text{Match}}$}. A pseudo-code implementation of this approach is presented in Algorithm 1, while more detailed information can be found in Appendix B.1.
\vspace{1mm}\\
\noindent \textbf{Overall Objective.} The full objective of UTSGAN is:
\begin{equation}
\begin{aligned}
\min_{{G,E}}{\max_{D}}\;\;
\mathcal{L}&_{\text{Real\_img}}  \\
+ & \lambda(\mathcal{L}_{\text{Recons\_img\_self}} +\mathcal{L}_{\text{Recons\_img\_cyc}}+ \gamma{\mathcal{L}_{\text{Recons\_out}}})\\
+ &\lambda_{\text{1}}\mathcal{L}_{\text{Recons\_trans}} \\
+ &\lambda_{\text{2}}\mathcal{L}_{\text{Recons\_newtrans}}  + \mathcal{L}_{\text{Real\_newtrans}}+\mathcal{L}_{\text{Real\_newimg}}\\
+ &\lambda_{\text{3}} (\mathcal{L}^{D}_{\text{Match}} +\mathcal{L}^{G}_{\text{Match}}) \label{eq:supervised_UTSGAN},
\end{aligned}
\end{equation}
\noindent where {\small$\lambda$}, {\small$\lambda_1$}, {\small$\lambda_2$} and {\small$\lambda_3$} control the relative importance of result consistency, transition consistency on observed translations, transition consistency on unobserved translations, and the generalization of overall consistency, respectively. Additionally, $\gamma$ is a hyperparameter controlling the importance of output reconstruction, with $\gamma=0$ when UTSGAN is trained with unpaired data.


\subsubsection{Training procedure}\label{sec:UTSGAN_model_training}
\begin{figure}[t]
	\centering	\includegraphics[width=0.48\textwidth]{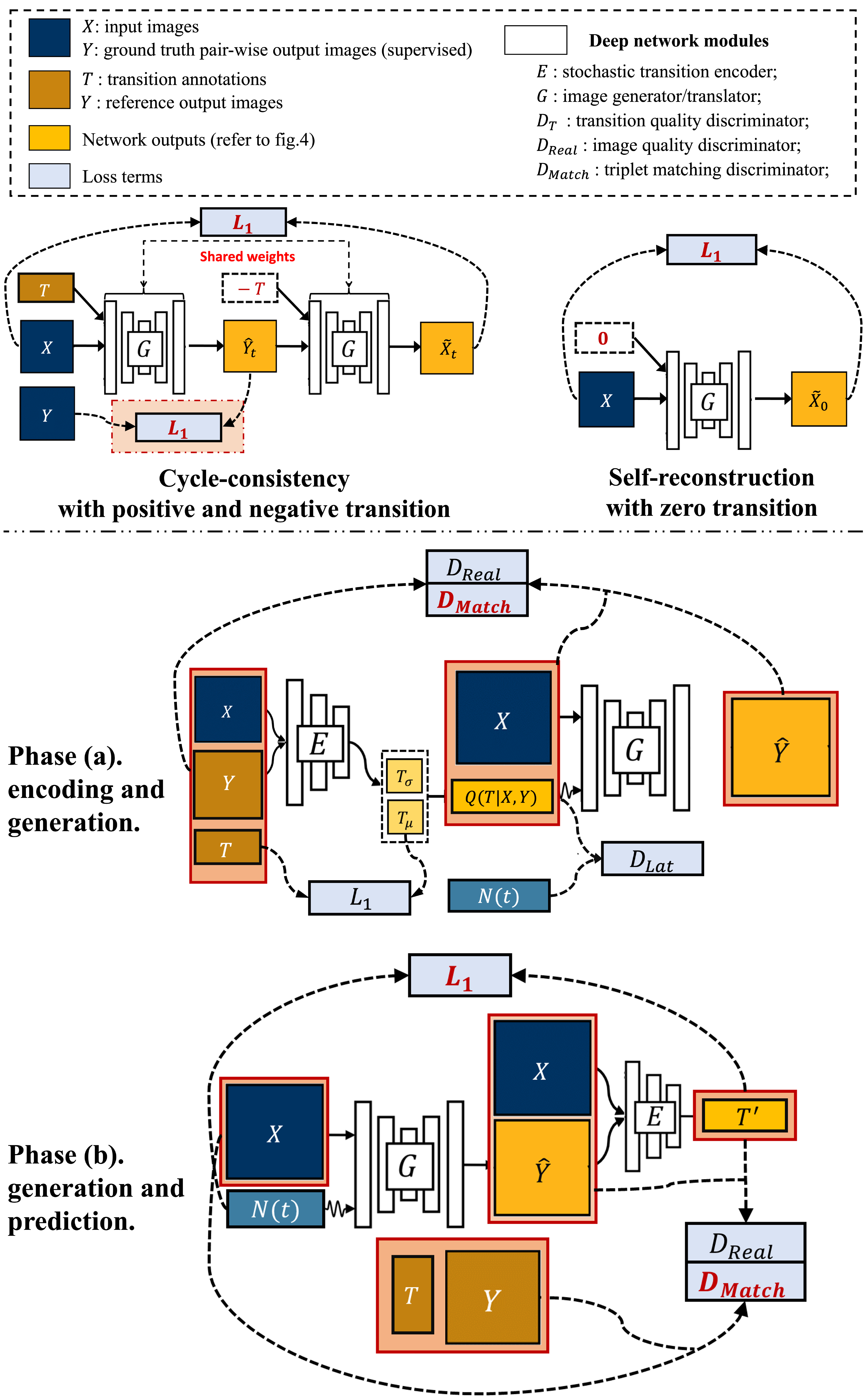}
	\vspace{-2mm}
	\caption{{The network design and training procedure of UTSGAN for supervised I2I translation.} If trained under an unsupervised setting, the $L_{1}$ loss for the reconstruction of the pair-wise output (marked orange in sub-figure. cycle consistency) will be dropped, \textit{i.e.}, $\gamma=0$ in Eq.~\eqref{eq:supervised_UTSGAN}. 
	}\label{fig:network_UTSGAN}
\end{figure}
\renewcommand{\algorithmicrequire}{\textbf{Function}}
\begin{algorithm}[t]
\begin{small}
		\caption{\label{alg:matching_algorithm}{Conditional adversarial loss for joint distribution matching}}
\begin{algorithmic}[1]
\REQUIRE $\text{Triplet\_Matching\_Loss} (x, t, y,t_{\times},y_{\times})$ 
\STATE ${\small s_{r} \leftarrow D_{\text{Match}} (x, t, y)\text{\COMMENT{real triplet}}}$ 
\STATE ${\small s_{f_1} \leftarrow D_{\text{Match}} (x, t, G(x,t))\COMMENT{\text{fake triplet}}}$
\STATE {\small$s_{f_2} \leftarrow D_{\text{Match}} (x, \hat{t}_{y}, G(x, \hat{t}_{y}))\COMMENT{\text{fake triplet: }\hat{t}_{y}=E(x,y)}$}\\ 
\STATE {\small$s_{f_3} \leftarrow D_{\text{Match}} (x, \tilde{t}, G(x, \tilde{t}))\COMMENT{\text{fake triplet: } \tilde{t}\sim q_{\phi}(t|x,y)}$}\\
\STATE {\small$s_{w_1} \leftarrow D_{\text{Match}} (x, t_{\times}, y )\text{\COMMENT{wrong triplet}}$}\\
\STATE {\small$s_{w_1} \leftarrow D_{\text{Match}} (x, t, y_{\times} )\text{\COMMENT{wrong triplet}}$}\vspace{1mm}\\
\STATE $\mathcal{L}^{D}_{\text{Match}} \leftarrow {(s_{r}-1)}^2 + \Sigma_{i=1}^{3}s_{f_i}^{2} + \Sigma_{i=1}^{2}s_{w_i}^{2} $\\
\STATE $L^{G}_{\text{Match}} \leftarrow {(s_{r}-1)}^2$\\
\RETURN $\mathcal{L}^{D}_{\text{Match}}$, $L^{G}_{\text{Match}}$
	\end{algorithmic}
\end{small}
\end{algorithm}

Fig.~\ref{fig:network_UTSGAN} illustrates the detailed network architecture and training procedure of UTSGAN in the supervised setting. The training process consists of two phases, where in the first phase (a), the generator $G$ and encoder $E$ are updated with their respective loss terms, while in the second phase (b), the focus is on training $G$ using the transition prediction loss terms, with $E$ being fixed. During both phases, the discriminators $D=\{D_{\text{Real}}, D_{T}, D_{\text{Match}}\}$  are iteratively trained with their corresponding loss terms. To enable bi-directional triplet matching, UTSGAN is trained using these two phases. A detailed description of the training process can be found in Appendix B.3.


\subsection{{Discussion}}\label{sec:discussion}

UTSGAN constructs a transition manifold $T$ and generalizes the regularization of transition consistency on on-manifold transitions. 
Compared to previous methods that solely regulate result consistency on the result manifold, our approach provides a tighter consistency for I2I with a more fine-grained and smoother consistency range.
Specifically, assuming our stochastic transition encoding network $E(x,y)$ contributes a Lipschitz continuous differentiable function {~\cite{qi2020loss}} over sample pairs on the result manifold with respect to the distance norm $\parallel\cdot\parallel$, we have:
\begin{small}
	\vspace{-1mm}\begin{equation}
		\parallel E(x,y)-E(x,\hat{y}) \parallel   \leq \kappa \parallel (x,y)-(x,\hat{y})\parallel \nonumber
		\vspace{-1mm}
	\end{equation}       
\end{small}
\noindent with a bounded constant $\kappa$. Thus, our transition consistency constraints in Eq.\eqref{eq:transition_consistency} and Eq.\eqref{eq:sample_consistent_t} lead to tighter consistency than the result consistency constraints in Eq.~\eqref{eq:result_consistency}. 
By generalizing such tighter consistency to the distribution-level, UTSGAN can achieve consistency for any on-manifold transitions and produce consistent results with smooth attribute changes.

{Our UTSGAN parameterizes the transition mapping with a function defined on the exact input-output image pair, which allows us to form three interrelated variables for I2I translation (Sec.3.1). 
We can holistically and coherently regulate the triplet consistency by constraining the result consistency and transition consistency through a bi-directional interactive process (Fig.~\ref{fig:utsgan_model}). Our stochastic transition encoder can instantiate various transitions with reasonable and highly probable unobserved attribute configurations. We can regulate transition consistency on translations triggered with these unseen transitions with coherent constraints as in the observed training translations. This coherent regularization of sample-level result consistency and transition consistency enables us to generalize triplet consistency to the distribution-level. UTSGAN's coherence in the regularization of consistency is based on the second-order logic of $X$ and $Y$~\cite{sep-logic-higher-order}, \textit{i.e.}, $X\overset{T}{\mapsto}Y$, which gives us superiority in model generalization regarding consistency.}

Note that among the previous methods, RelGAN is the most relevant work that is similar to our UTSGAN. 
However, to facilitate model generalization, RelGAN models continuous transitions using a linear coefficient $\alpha$ on predefined transitions, \textit{i.e.}, $\tilde{t}=\alpha\cdot t$. This is a simple linear and deterministic operation among the various stochastic operations defined by our transition encoding module. 
RelGAN's linear operation can only model a fixed set of linearly operated variants of the observed $t$ on a predefined linear plane, making it unable to cover the massive unobserved attribute configurations during training. 
Furthermore, RelGAN enforces consistency for interpolated translations on $\alpha$ instead of on the transition $\tilde{t}$ itself. 
This regularization diverges from the result consistency regularizers imposed for the forward transition conditional generation process, hindering the model from achieving a coherent regularization of consistency for generalization.

\begin{figure*}[t]
	\centering
	\includegraphics[width=0.98\textwidth]{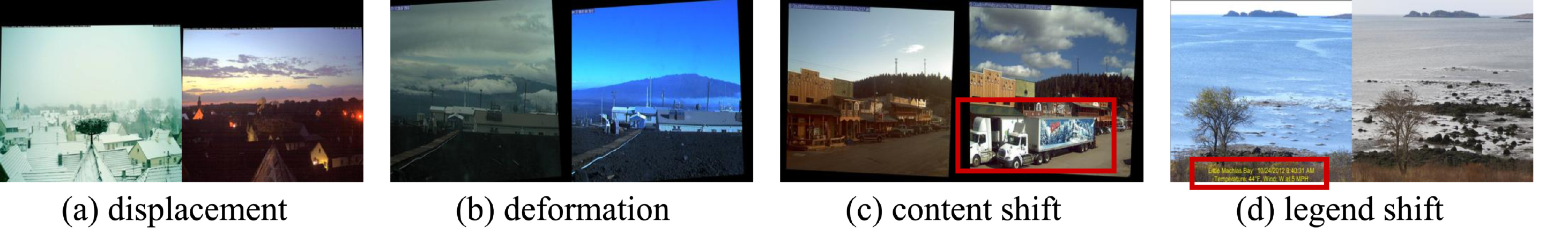}
	\vspace{-3mm}%
	\caption{Example paired data in the transient attribute dataset. 
	}\label{fig:night2day_dataset}
	\vspace{-4mm}
\end{figure*}

\section{Experiments}\label{sec:experiments}

In this section, we will describe the experimental settings in Section~\ref{sec:exp_setting}, followed by the comparison results on four representative I2I tasks: outdoor scene editing in Section~\ref{sec:exp_outdoor}, face editing in Section~\ref{sec:exp_face_editing}, multi-domain style transfer and generalization in Section~\ref{sec:exp_style_transfer}, and image inpainting in Section~\ref{exp:image_inpainting}. 
We will use five datasets for our experiments.

\subsection{Experimental settings}\label{sec:exp_setting}

\noindent \textbf{Network architectures.} 
In our experiment, the network configuration of UTSGAN is as follows:
\begin{itemize} \setlength{\itemsep}{1mm}
    \item Generator $G$: we adopt a U-NET structure defined in~\cite{DBLP:conf/miccai/RonnebergerFB15};
    
    \item Transition encoder $E$: we utilize a ResNet consisted of four residual blocks~\cite{He2015} to construct $E$;
    
    \item Discriminator on transition $D_{T}$: we build {a four layers-MLP};
    
    \item Other discriminators {\small$D = \{D_{\text{Real}}, D_{\text{Match}}\}$}, have a shared feature sub-network comprised of six convolutional layers with a stride of 2. Each discriminator has its output layers added onto the feature sub-network. 
\end{itemize}

\noindent We present configuration details of each module in {Appendix~B.2}.

\noindent \textbf{Training details.} Our UTSGAN model was trained using an Adam optimizer~\cite{DBLP:journals/corr/KingmaB14} with hyperparameters $\beta_{1}=0.5$, $\beta_{2}=0.999$, a batch size of 32, and a learning rate of 0.00006. The loss coefficients for Eq.~\eqref{eq:supervised_UTSGAN} were set to $\lambda=10$, $\lambda_{1}=1$, $\lambda_{2}=0.5$, and $\lambda_{3}=1$. 
The coefficient $\gamma$, which represents the relative importance of output reconstruction, varied depending on the task: for unsupervised tasks, such as face editing and multi-domain style transfer, we set $\gamma=0$, while for supervised tasks, such as outdoor editing and inpainting, we set $\gamma=10$ for outdoor editing and $\gamma=25$ for image inpainting.


We emphasize that our work focuses on achieving consistent I2I translation with transition consistency. 
As a result, we adopt basic network structures to highlight the benefits derived from the generalization of consistency. 
However, more advanced network structures, such as pre-trained generators like BigGAN{\cite{brock2018large}} or StyleGAN {\cite{karras2019style,karras2020analyzing,karras2021alias}}, can also be incorporated into UTSGAN to improve result quality with high-fidelity generation.



\subsubsection{Outdoor scenes editing}

We conduct outdoor scenes editing tasks based on {Transient Attributes Dataset}~\cite{laffont2014transient}. This task is set with a supervised setting, in which the transition $t$ characterizes semantic attribute changes through each translation, \textit{i.e.}, {\small$x\overset{t = z_y-z_x}{\longrightarrow}{y}$}, where $z_{y}$ and $z_{x}$ are attribute annotations of the output image $y$ and the input image $x$. 

\noindent \textbf{{Transient attribute dataset:}} This dataset contains $8,571$ scene images taken from $101$ outdoor webcams. Those images are captured over long periods, exhibiting drastic changes in appearance. Each of these images is annotated with $40$ transient attributes, such as ``sunny'', ``bright'', and ``ice''.

We randomly select one image in the same scene as the output for each image, thus constructing $8571\times1$ paired images for this task. 
$8071$ of these images are randomly selected for training, and the rest of the data are for testing.
Note that, since these images are captured in natural scenes, there exist various irregular shifts between the paired images for model training, \textit{e.g.}, displacement, deformation, content shift, and legend shift.
We present example data pairs within the dataset in Fig.~\ref{fig:night2day_dataset}. 

\noindent \textbf{Baselines:} We compare UTSGAN with three state-of-the-art general I2I translation frameworks, including
Pix2Pix~\cite{DBLP:conf/cvpr/IsolaZZE17}, 
BicycleGAN~\cite{DBLP:conf/nips/ZhuZPDEWS17},
and AugCGAN~\cite{DBLP:conf/icml/AlmahairiRSBC18},
in this task.


\subsubsection{Face editing}

We conduct face editing on the CelebA-HQ dataset under an unsupervised setting. In this task, the transition $t$ specifies the change of facial attributes during each translation, \textit{i.e.}, {\small $x\overset{t = z_y-z_x}{\longrightarrow}{y}$}. 

\noindent {\textbf{CelebA-HQ dataset}}:
This dataset is a high-quality version of the CelebFaces Attributes (CelebA) dataset~\cite{DBLP:conf/iclr/KarrasALL18}. 
It contains $30,000$ high-resolution face images with a resolution of {$\small1024 \times 1024$}. 
Each image is annotated with $40$ binary attributes such as hair color, gender, and age. 
For the face editing task, we center-crop and resize the images to $256 \times 256$ and select $10$ identifiable attributes, including hair color (``black'', ``blond'', ``brown''), dress-up (``eyeglasses'', ``bangs'', ``mustache'', ``pale skin'', ``smiling''), gender (``male''/ ``female''), and age (``young''/ ``old''). 
We use a 90/10 split of the dataset for training and testing.


\noindent \textbf{Baselines:} We compare UTSGAN with four state-of-the-art face editing methods: AttGAN~\cite{DBLP:journals/tip/HeZKSC19},
StarGAN~\cite{DBLP:conf/cvpr/ChoiCKH0C18},
RelGAN~\cite{DBLP:conf/iccv/LinWCCL19}
and StarGANv2~\cite{choi2020stargan}, in this task.


\subsubsection{Multi-domain style transfer}

Our UTSGAN is handy to conduct I2I translation tasks with multiple target domains {\small $Y^{\{1,2,...n\}}$}, \textit{i.e.}, multi-domain I2I translation, with $t$ specifies the target domain index of each transformation, \textit{i.e.}, {\small $x\overset{t}{\longrightarrow}{y^{t}}$}. 
Here, we apply UTSGAN to a multi-domain style transfer {on two datasets} to testify its performance in such tasks. 

\noindent {\textbf{Photo2Art dataset:}} 
We constructed the Photo2Art dataset by combining four commonly used photo-to-artistic painting datasets for multi-domain style transfer, namely, \textit{photo$\rightarrow$cazanne}.
\textit{photo$\rightarrow$monet}, \textit{photo$\rightarrow$ukiyoe}, and \textit{photo$\rightarrow$vangogh}. 
For each photo image $x$, we randomly sampled one art painting as $y$, resulting in a triplet data sample where $t$ specifies the style index of the output. 
The dataset contains a total of $25,148$ photo-to-art image pairs. 
We randomly selected $500$ samples from the entire dataset for testing and used the remaining data for training.


%
\noindent {\textbf{AFHQ dataset:} Animal FacesHQ (AFHQ) is a dataset consisting of $15,000$ high-quality animal face images from three domains: ``{cat}'', ``{dog}'', and ``{wildlife}''. Each domain contains diverse images of various breeds (more than eight per domain), making the multi-domain style transfer problem more challenging. For example, the \textit{wildlife} domain includes face images of lions, tigers, wolves, and foxes (see Fig.~\ref{fig:afhq}). All images are resized to 256 $\times$ 256, and we adopt a 90/10 split of these images for training and testing.}


\noindent \textbf{Baselines:} We compare UTSGAN with three multi-domain translation methods here: MUNIT~\cite{huang2018munit}
, {DLOW}~\cite{DBLP:conf/cvpr/GongLCG19}, {and DRIT++~\cite{lee2020drit++}}.
For MUNIT, we mixed the images of all four styles as the target domain and set the dimension of its style code {\small$d_{s}=4$}.


\begin{figure*}[t]
	\centering
	\includegraphics[width=0.91\textwidth]{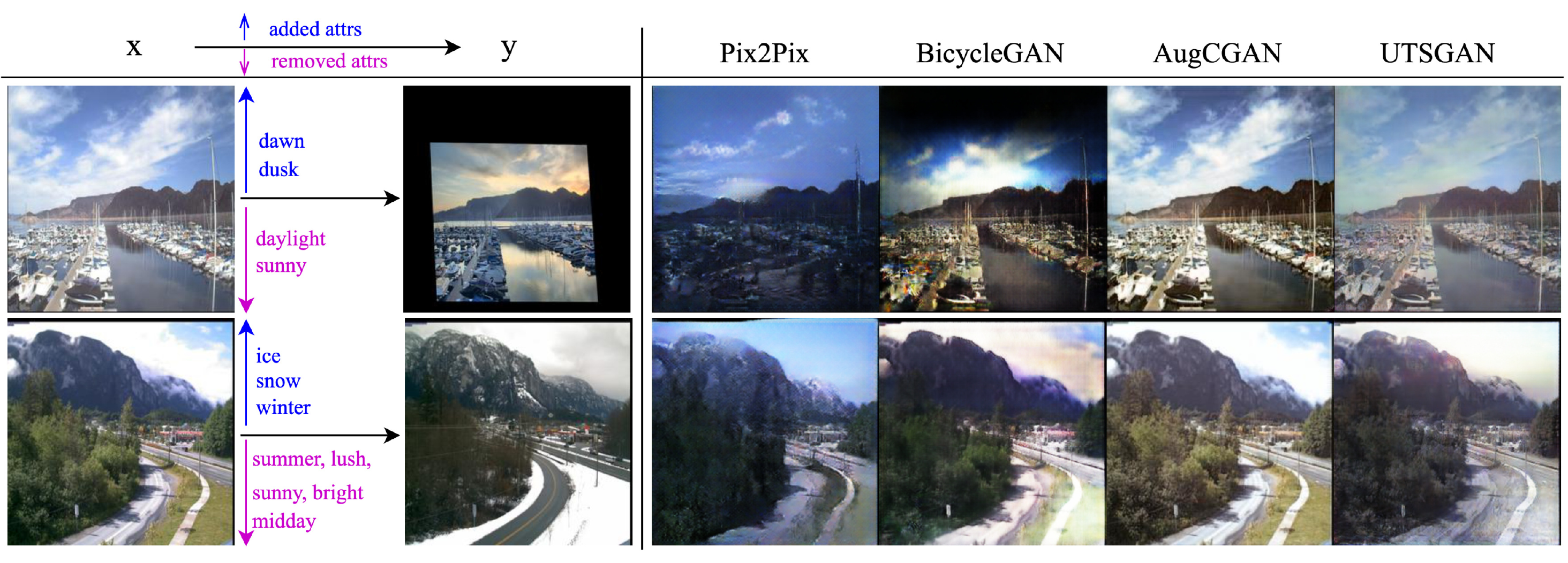}
	\vspace{-3mm}
	\caption{{Comparisons of UTSGAN with {Pix2Pix~\cite{DBLP:conf/cvpr/IsolaZZE17},  BicycleGAN~\cite{DBLP:conf/nips/ZhuZPDEWS17}} and AugCGAN~\cite{DBLP:conf/icml/AlmahairiRSBC18} on supervised outdoor scenes editing task.} The attributes in blue are desired to be added to the output image, while the attributes in red are to be removed from the input image. 
    Please zoom in for details. 
	}\label{fig:night2day_a2b}
	\vspace{-4mm}
\end{figure*}
\begin{figure*}[t]
	\centering
	\includegraphics[width=0.8\textwidth]{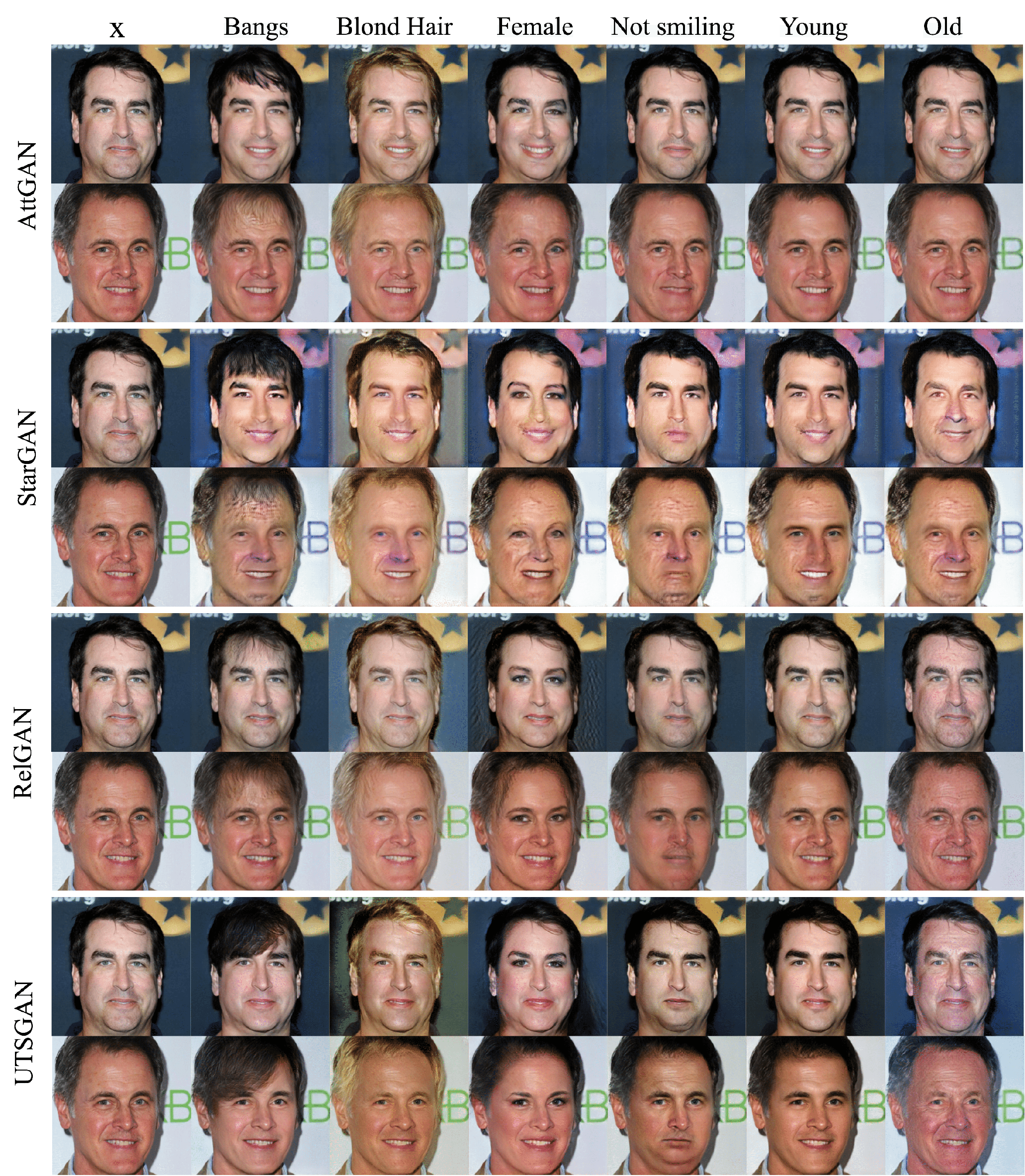} 
	\vspace{-2mm}
	\caption{Comparison on single attribute editing among UTSGAN, AttGAN, StarGAN and RelGAN on CelebA-HQ dataset. Please zoom in for details. 
	}\label{fig:celeba_binary_editing} 
	\vspace{-4mm}
\end{figure*}
\subsubsection{Image inpainting}\label{sec:setting_inpainting}
We apply UTSGAN to image inpainting tasks, where $t$ specifies the location of the masked regions, conveying the relative position change between the realistic image region and the missing region in each masked input. 
The translated results should be realistic and plausible to fill the missing regions. 
More importantly, UTSGAN's results are expected to have less evident boundaries since we enforced transition consistency between the masked region and the in-painted pixels region in the generated outputs. 
It is worth noting that since we define $t$ with quantitative relative locations here, we focus on evaluating UTSGAN with regular types of masking, {\textit{e.g.} squared marks, where the marked region can be easily represented with its relative left-top location and size of the mask.}  
{You can also define $t$ with the physical mask or its positional embedding to instantiate UTSGAN to cope with regular masks.}  

{To clarify the semantics of the transition in the inpainting task, we extend the $t$ variable with an additional dimension to specify the translation direction in UTSGAN, following other generative inpainting methods~\cite{iizuka2017globally,DBLP:conf/cvpr/Yu0YSLH18}. In this way, $t$ contains the dimensions that specify the location of the masked regions, as well as the translation direction. 
Specifically, $+1$ indicates the positive direction $x\rightarrow y$, where the masked images are inpainted to complete ones, while $-1$ means the inverse direction, $y\rightarrow x$, where the complete image is masked to create the corresponding incomplete image.}

 

\noindent {\textbf{CelebA-HQ dataset for inpainting:}} We resize the original images in CelebA-HQ, which are of size $1024 \times 1024$, to $256 \times 256$, and randomly erase a $128 \times 128$ squared region from each image at a randomly sampled spatial location. 
In the context of UTSGAN, $x$ represents the masked incomplete image, $y$ represents the original complete image, and $t$ specifies the masked region in each data sample. 
We randomly select $1,000$ samples from the dataset for testing and use the remaining data for training.


\noindent {\textbf{FFHQ dataset:} Flickr-Faces-HQ (FFHQ) is a benchmark dataset with 70,000 high-quality human face images with wide variation in gender, age and image background~\cite{karras2019style}. 
For our experiment, we center-crop and resize all original images to $256 \times 256$ and randomly mask a $128 \times 128$ squared region. 
We use a $90/10$ split of these images for training and testing.
}



\noindent \textbf{Baselines:} We compare UTSGAN with three state-of-the-art image inpainting methods in this task, including  CA-Inpainting~\cite{DBLP:conf/cvpr/Yu0YSLH18},
PIC-Net~\cite{zheng2019pluralistic}
and PEN-NET~\cite{yan2019PENnet}.

\subsection{Experimental results}\label{sec:exp_results}

\subsubsection{{Outdoor scenes editing}}\label{sec:exp_outdoor}
\begin{table}[t]
\begin{center}
			\renewcommand{\arraystretch}{1.1}
			\caption{Comparison of the SSIM($\uparrow$) and PSNR($\uparrow$) scores of the translated results of each method in outdoor scenes editing. The best results are highlighted in bold. The larger the better.}
			\vspace{-2mm}
			\label{tab:ssim_night2day}
			\setlength{\tabcolsep}{0.9mm}{
				\scalebox{0.98}{
\begin{tabular}{c|c|cccc}
\toprule
{Metrics}      & {Image Pairs}             & {Pix2Pix}   & {BicycleGAN}  & {AugCGAN}  & {UTSGAN\textit{(ours)}} \\ \hline
\multirow{3}{*}{SSIM} & $(x,\hat{x})$  &    -               &     0.52        &     0.09 &        \textbf{0.82}     \\
                      & $(y,\hat{y})$  &    0.25            &     0.42         &    0.07    &        \textbf{0.75}     \\ 
                    \hline
\multirow{3}{*}{PSNR} & $(x,\hat{x})$  &    -               &     14.05        &    2.32 &        \textbf{22.52}     \\
                      & $(y,\hat{y})$  &    8.5             &     11.64        &    1.98 &        \textbf{19.10}     \\
\bottomrule
\end{tabular}}}
\end{center}
\vspace{-4mm}
\end{table}
\begin{table}[tbp]
\begin{center}
			\renewcommand{\arraystretch}{1.2}
			\caption{{Comparison of the LPIPS diversity score ($\uparrow$) for \textbf{multi-output translation} of each method in outdoor scenes editing.} 
}    			\vspace{-2mm}       \label{tab:diversity}
			\setlength{\tabcolsep}{1.2mm}{
				\scalebox{0.98}{
\begin{tabular}{c|c|c|c|c}
\hline 
\toprule
         & Pix2Pix   & {BicycleGAN} & {AugCGAN } & {UTSGAN } \\ \hline
{LPIPS} & -& $0.0400\pm0.017$  & $0.0997\pm0.0288$  & $0.0982\pm 0.0273$\\
\bottomrule
\end{tabular}
}}
\vspace{-6mm}
\end{center}
\end{table}
\noindent \textbf{Visual quality comparison:} Fig.~\ref{fig:night2day_a2b} presents example results of these methods. 
In these results, our UTSGAN outperforms all three methods, with the translated results achieving better quality and presenting distinguishable desired attributes. Specifically, in the first row, Pix2Pix misses the water region existing in $x$; the result of BicycleGAN is influenced by deformation in $y$, with the details around the border of the image missing; AugCGAN fails to perform a translation with the desired property, generating a brighter instead of a ``\textit{darker}'' image. 
In the second row, we expect the translated images to be a scene in ``\textit{winter}'' with ``\textit{ice}'' and ``\textit{snow}''. 
However, compared to UTSGAN, results from Pix2Pix, BicycleGAN, and AugCGAN all show an increased amount of green areas. 
More qualitative result comparisons are given in {Appendix~C}.

\noindent \textbf{Quantitative results:}
 We report average SSIM~\cite{bulat2018super}, PSNR~\cite{Zhang_2018_ECCV}, and LPIPS distance, to quantitatively compare UTSGAN with the three methods. SSIM and PSNR are used to evaluate the generated image quality, LPIPS distance is used to evaluate result diversity.

In Table.~\ref{tab:ssim_night2day}, we report the average SSIM and PSNR scores of the generated images, including the reconstruction of the input ($\hat{x}$) and the generation result ($\hat{y}$), of each method. {We find that our UTSGAN achieves the highest score.} Since AugCGAN mainly emphasizes many-to-many translation, its generated results present low similarity with respect to its example output. 

In Table.~\ref{tab:diversity}, we report the LPIPS distance of random generation results of BicycleGAN, AugCGAN, and UTSGAN. We average the distance between each input image and its generated results and report the average diversity score of all images. We can see UTSGAN significantly outperforms BicycleGAN. AugCGAN gets a score comparable to UTSGAN; however, its results fail to present desired attributes as ours, as shown and discussed in Fig.~\ref{fig:night2day_a2b}. 



\vspace{-4mm}
\subsubsection{{Face editing}}~\label{sec:exp_face_editing}
We investigate the face editing task with three different settings:\begin{enumerate*}
    \item[{\textit{(\romannumeral1)}}] single attribute editing; 
    \item[{\textit{(\romannumeral2)}}] single attribute interpolation; and 
    \item[{\textit{(\romannumeral3)}}] multiple attributes editing. 
\end{enumerate*} 

\vspace{1mm}
\noindent {\textbf{{\textit{(\romannumeral1). }} Single attribute editing}}
\begin{figure*}[t]
	\centering
 	\includegraphics[width=0.94\textwidth]{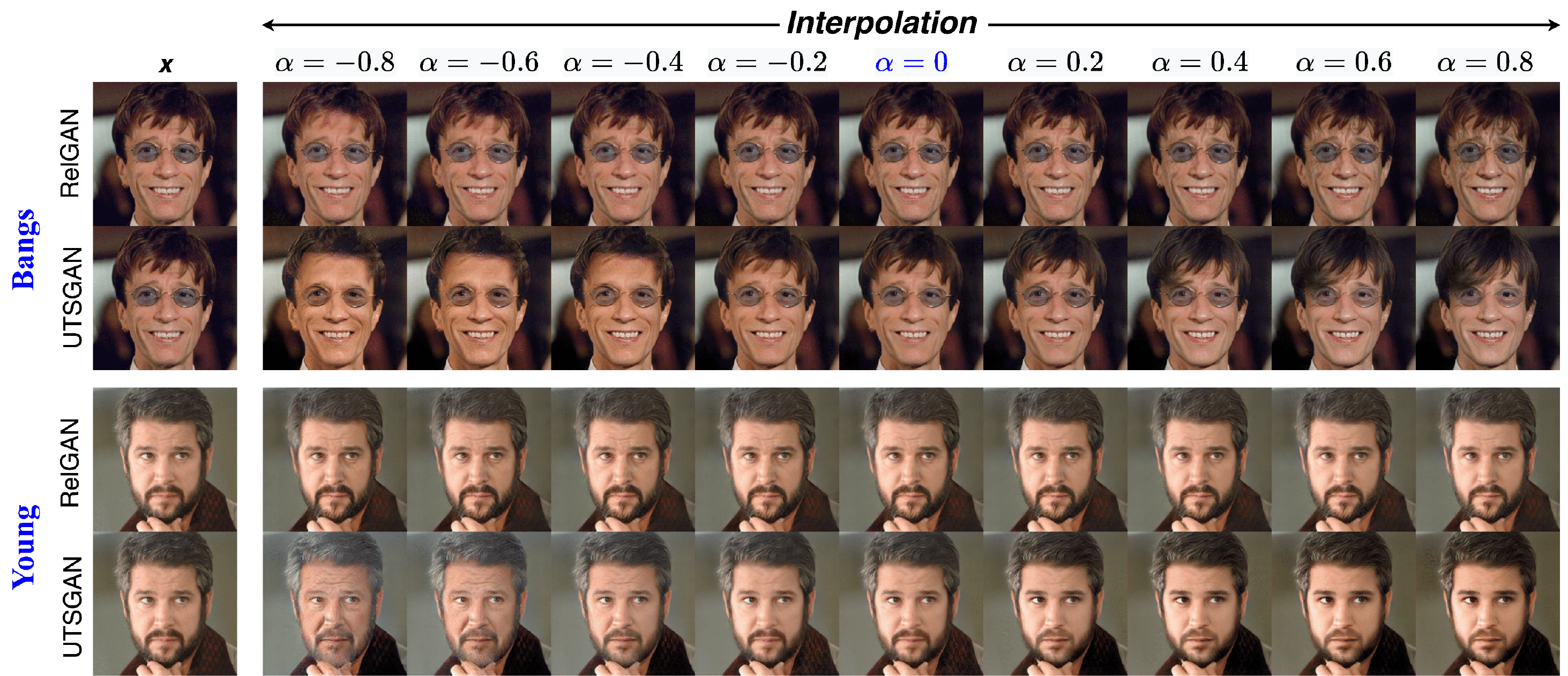}
	\vspace{-2mm}
	\caption{Comparison of RelGAN and UTSGAN on single attribute interpolation. 
		Compared with RelGAN, the results of UTSGAN present more evident changes in a relatively longer range. Please zoom in for details.} \label{fig:supp_celeba_interpol}
	\vspace{-4mm}
\end{figure*}
\begin{figure*}[t]
	\centering
	\includegraphics[width=0.91\textwidth]{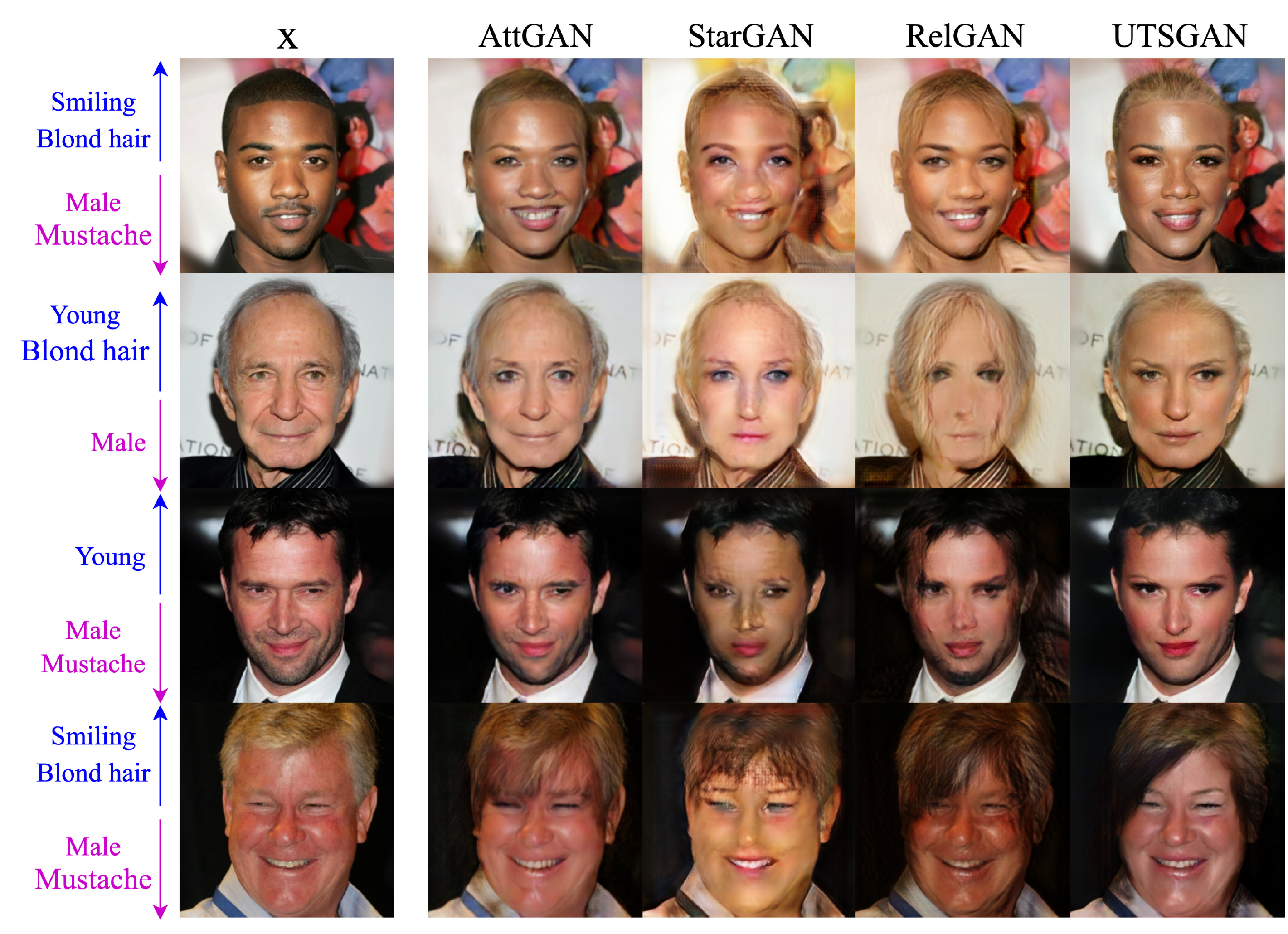}
	\vspace{-4mm}
		\caption{Comparison results of \textit{multiple} facial attribute editing among these methods. 
		The attributes in blue are desired to be added, while the attributes in red are to be removed from the input image. Please zoom in for details. 
		} \label{fig:celeba_multi_editing}
		\vspace{-4mm}
\end{figure*}
\begin{figure*}[t]
	\centering
 	\includegraphics[width=0.91\textwidth]{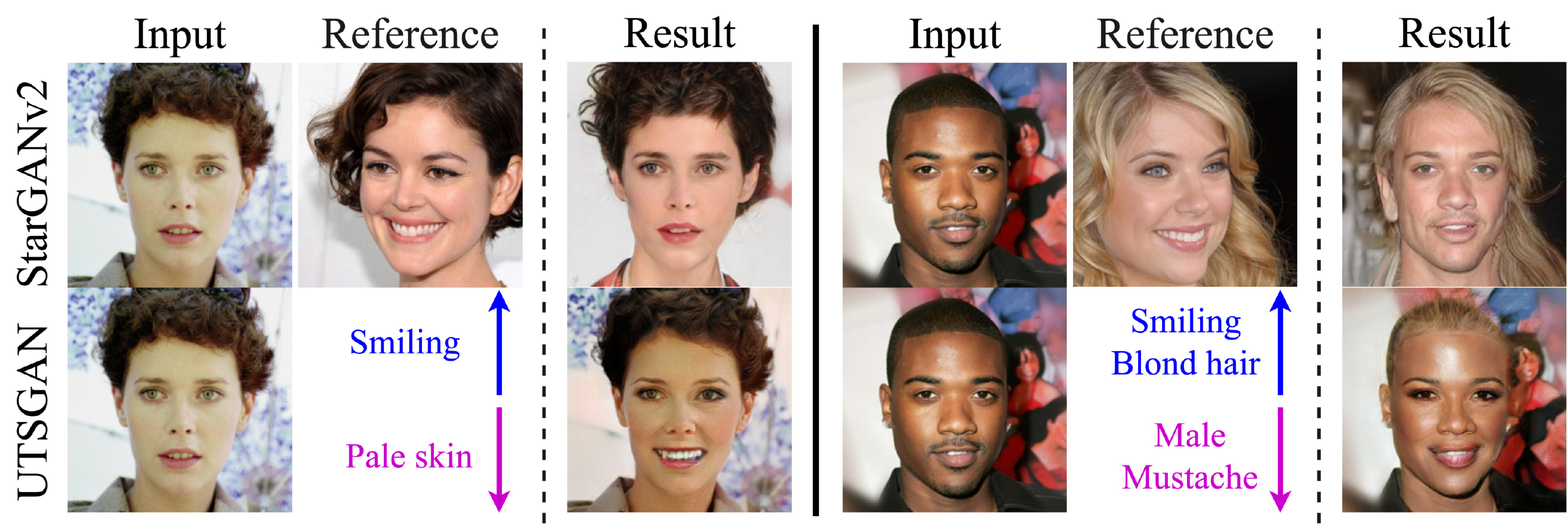}
	\vspace{-4mm}
	\caption{Example results of UTSGAN and StarGANv2. Our UTSGAN subtly edits a human face making it present desired semantic attributes, while StarGANv2 changes the overall style of the input human face images to the style of the reference image. Please zoom in for details. 
	} \label{fig:celeba_starv2}
	\vspace{-4mm}
\end{figure*}
We first evaluate UTSGAN and the baseline methods on the commonly studied translation task with single attribute change. 

\noindent \textbf{Visual quality comparison:}  
Fig.~\ref{fig:celeba_binary_editing} shows example results of UTSGAN and the baseline methods.
Generally, the results of UTSGAN are the most satisfactory, with high-quality translated images showing evident desired attribute changes. 
The results of StarGAN are the worst, with notable artifacts in most images. 
The translated results of AttGAN are not reasonable, as most of its results trigger the smiling attribute (mouth changes to be open), which is not the desired attribute. 
For RelGAN, some of its results present unexpected blurriness, \textit{e.g.,} its ``\textit{Female}'' editing results in the first set and ``\textit{Smiling}'' editing results in the second set. 
In addition, its editing results on some attributes are not satisfactory, as they contain obvious artificial traces according to human perception. 
For example, in the second set, the editing results of ``\textit{Blond}'' hair color and ``\textit{Female}'' both present factitious hairs on the forehead of the human face. 
In contrast, the translated results of UTSGAN have no visible quality flaws and present evident editing on each attribute. 
Two result improvements are worth highlighting: First, UTSGAN can change the hairstyle of the human face when editing an input to be ``\textit{Female}''. This is because our transition encoding captures the relations among different attributes. 
Second, in terms of the editing results on ``\textit{Young}'' and ``\textit{Old},'' the results of our UTSGAN present more evident visual contrast, especially in the color of the hair.


\vspace{1mm}
\noindent \textbf{{\textit{(\romannumeral2).} {Single attribute interpolation}}}

We specifically compare UTSGAN with RelGAN on single attribute interpolation to verify UTSGAN's ability to perform translations with subtle attribute changes. We conduct attribute interpolation on some visually legible attributes, {\textit{e.g.}, ``{Bangs}'', ``{Old}'' and ``{Young}''.} 

\noindent \textbf{Visual quality comparison:} 
Fig.~\ref{fig:supp_celeba_interpol} presents example results, in which UTSGAN exhibits smooth-varying interpolations for each attribute. 
Compared to RelGAN, the interpolation results of UTSGAN are superior in presenting more evident attribute changes. 
For instance, in the interpolation of ``\textit{Bangs}'', RelGAN's results fail to exhibit distinguishable variations. 
This is because UTSGAN learns a manifold for $t$, which is the transition manifold, instead of interpolating on the fixed attribute range, \textit{i.e.}, the linear plane of {\small$t = z_{b}-z_{a}$}. 
Consequently, this results in a longer span that is induced by the intrinsic property of the attributes. 
More importantly, by modelling the transition manifold, UTSGAN can present more smooth and evident editing on generated results along with the interpolation of attributes than RelGAN.

\vspace{1mm}
\noindent {\textbf{{\textit{(\romannumeral3).}} Multiple attributes editing}}

We highlight the comparison on this multi-attributes face editing task, where the translations convey complex attribute combinations that are highly possible to be unseen during training.

\noindent \textbf{Visual quality comparison:}
We present example results in Fig.~\ref{fig:celeba_multi_editing}. 
The results show that our UTSGAN generates more realistic and semantically reasonable results than previous methods. 
Specifically, the results of StarGAN present clear deficiencies, especially 
for the test case in the first row. 
We believe it is due to the mismatch between the domain classifier and attributes annotation in StarGAN. 
RelGAN obtains blurry results in this setting since it simply models the unseen transitions with linear operations, \textit{i.e.}, {\small$\tilde{t}=\alpha\cdot{t}$}, which can not cover other flexible combinations of attributes that do not present in the training set. 
For AttGAN, although its results are more realistic than RelGAN and StarGAN, they fail to present meaningful and harmonious editing on multiple attributes, \textit{e.g.}, ``\textit{Moustache}'' still exists when we change the gender from ``\textit{Male}'' to ``\textit{Female}'' (See the second image in the third row). This is because AttGAN treats each attribute individually and fails to model the relationships among different attributes. 
We present more results in {Appendix~C}.

\noindent \textbf{Quantitative results:}
We qualitatively evaluate the multiple attribute editing results of the methods in two aspects:
\begin{enumerate*}
    \item[{\textit{(\romannumeral1)}}] quality of the generated images;
    \item[{\textit{(\romannumeral2)}}] attribute classification accuracy. 
\end{enumerate*}

\noindent {\textbf{{\textit{(\romannumeral1).}} Quality of the generated images}} Table.~\ref{tab:multi_attrs_fid} presents the comparison regarding Fr\'echet Inception Distance {(FID)}~\cite{heusel2017gans} (lower is better) between the input image $x$ and the multiple attributes translated results. We can see UTSGAN achieves the best results among all the methods, indicating its results are of high quality. 
\begin{table}[t]
\begin{center}
		\renewcommand{\arraystretch}{1.2}
			\caption{The comparison of Fr\'echet Inception
Distance (FID $\downarrow$) between input images and {\textit{\textbf{multiple attribute editing}}} results of each method. A lower value means better. 
}           \label{tab:multi_attrs_fid}
			\vspace{-2mm}
			\setlength{\tabcolsep}{1mm}{
				\scalebox{1.2}{
\begin{tabular}{c|cccc}
\hline 
\toprule
    &AttGAN & StarGAN & RelGAN  &UTSGAN  \\ \hline
FID  & 72.22 & 125.94  & 71.78   & \textbf{66.93} \\ 
\bottomrule
\end{tabular}
}} \vspace{-3mm}
\end{center}
\end{table}
\begin{table}[t]
	\begin{center}
		\renewcommand{\arraystretch}{1.2}
		\caption{Classification accuracy ($\uparrow$) of \textit{\textbf{single and multiple attribute editing results}} on CelebA-HQ.} \label{tab:multi_attrs_acc}
		\vspace{-2mm}
		\setlength{\tabcolsep}{1.2mm}{
			\scalebox{0.936}{
				\begin{tabular}{c|c|cc|cc} 
					\toprule
					\multirow{2}{*}{\textbf{Attributes}}&\cellcolor{mygray3}&\multicolumn{2}{c|}{{\textbf{Single attribute editing}}} & \multicolumn{2}{c}{{{\textbf{Multiple attribute editing}}}} \\ \cline{3-6}
					& \multirow{-2}{*}{\cellcolor{mygray3}\scriptsize{\textbf{{CelebAHQ}}}}& {{RelGAN}}&{{UTSGAN}}&{{RelGAN}}&{{UTSGAN}}\\ \hline
					\textit{Bangs}      				&\cellcolor{mygray3}\scriptsize{95.47}&  92.60    & \textbf{97.70}        &  92.80          &  \textbf{95.70}\\
					\textit{Black\_hair}			  &\cellcolor{mygray3}\scriptsize{90.93}&  78.60    & \textbf{85.90}       &  \textbf{90.00} &  88.60         \\
					\textit{Male}       				  &\cellcolor{mygray3}\scriptsize{98.23}&  95.23    & \textbf{96.53 }      &  93.90          &  \textbf{ 95.60} \\
					\textit{Mustache}   		 	 &\cellcolor{mygray3}\scriptsize{96.63}&  58.73    & \textbf{63.23}        &  91.35          &  \textbf{97.25}\\
					\textit{Smiling}    				&\cellcolor{mygray3}\scriptsize{94.47}&  93.60    & \textbf{98.45}       &  \textbf{92.15} &  91.80     \\ \hline
					\textit{\textbf{Average}}	 &\cellcolor{mygray3}\scriptsize{95.07}	& 83.75  &  \textbf{88.36}  & 92.04 & \textbf{93.79}  \\ 
					\bottomrule
				\end{tabular}
			}
		} 
	\end{center}
	\vspace{-5mm}
\end{table}
\noindent {\textbf{{\textit{(\romannumeral2).}} Attribute classification accuracy.}} To evaluate the consistency between the translated results and the desired attribute, we conduct an experiment to predict the attributes of the generated images. We train $5$ binary classifiers (ResNet18~\cite{He2015}) for the five most distinguishable attributes, \textit{i.e.}, ``{Bangs}'', ``{Black\_hair}'', ``{Male}'', ``{Mustache}'', ``{Smiling}'', on the CelebAHQ dataset, and apply them to the generated images of RelGAN and UTSGAN. The classification results are presented in Table~\ref{tab:multi_attrs_acc}. It demonstrates that our UTSGAN surpasses RelGAN in achieving semantically consistent translated results in the face editing task. 
\begin{figure*}[t]
	\centering
 	\includegraphics[width=0.96\textwidth]{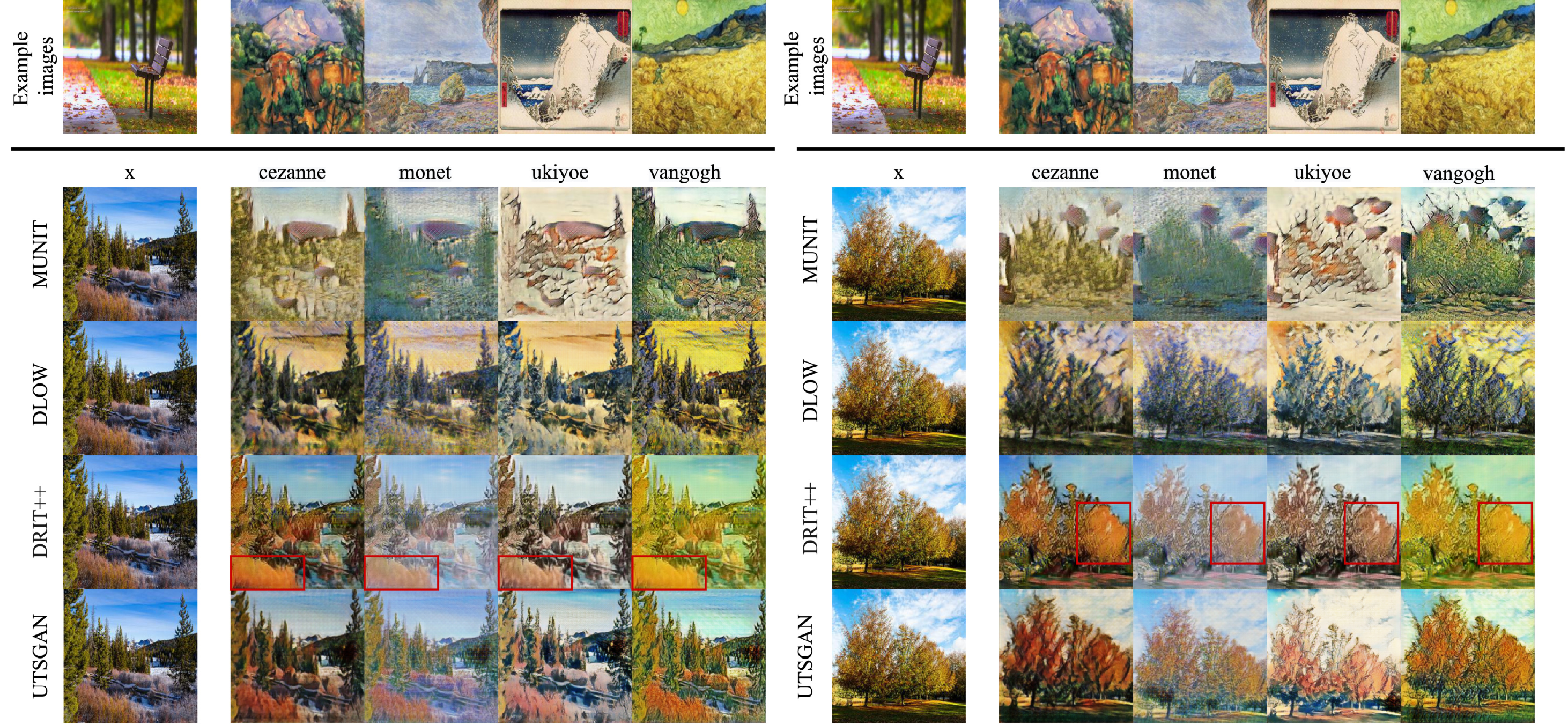}
\vspace{-2mm}
	\caption{Example results of MUNIT, DLOW, DRIT ++ and UTSGAN on multi-domain style transfer on the Photo2Art dataset. Zoom in for details. 
	} \label{fig:styletranfer_editing}
\vspace{-4mm}
\end{figure*}
\begin{figure*}[t]
	\centering
 \includegraphics[width=0.96\textwidth]{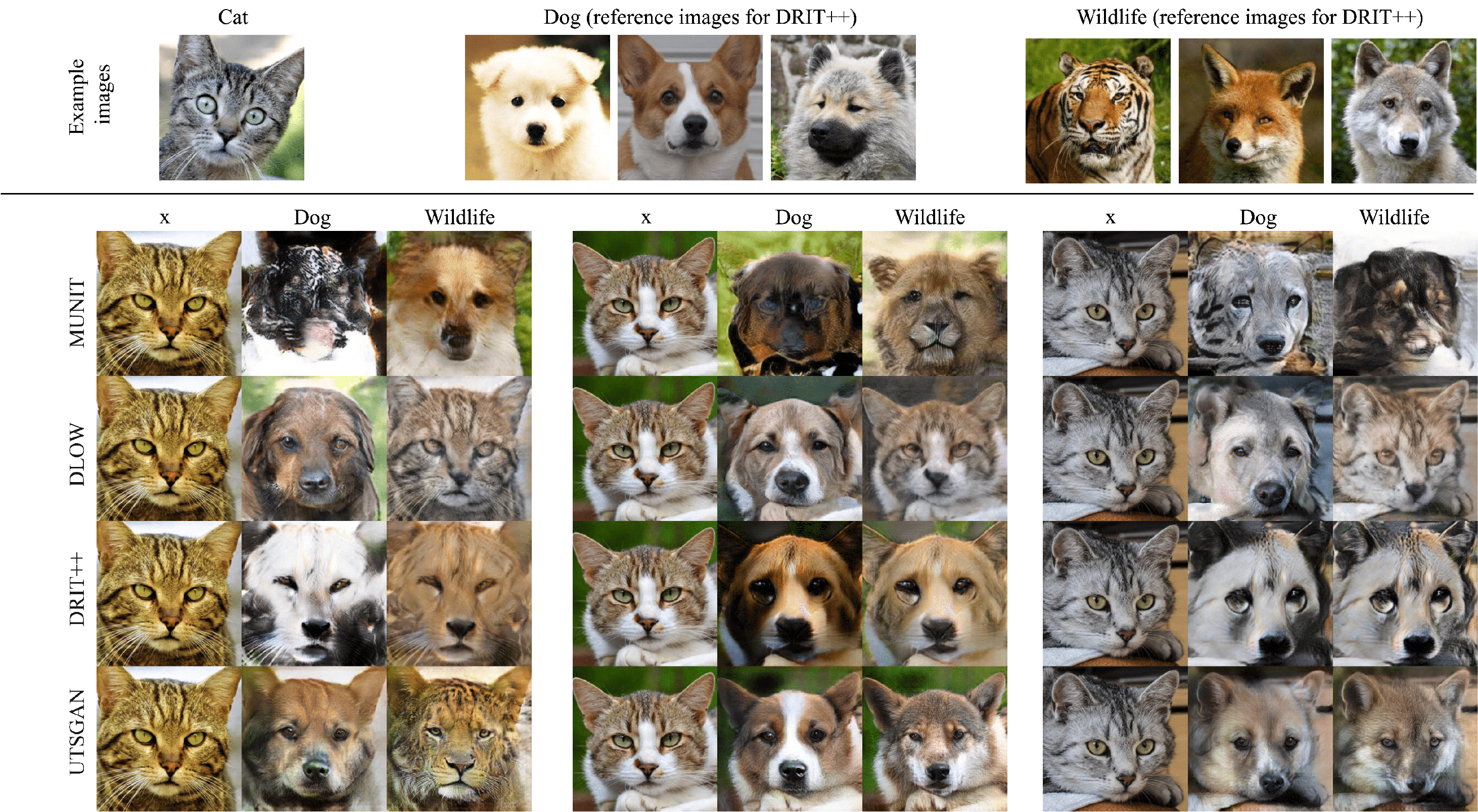}
	\vspace{-2mm}
	\caption{{Example results of MUNIT, DLOW , DRIT++, and UTSGAN on multi-domain translation on the AFHQ dataset. Please zoom in for details. Note that, different from the other method that {only requires the target domain index}, DRIT++, requires example images of the target domain as input to perform reference-guided multi-domain translation. The example images of ``'\textit{Dog}'' and ``\textit{Wildlife}'' shown in the first row are also the reference images of DRIT++ for each set of experiments. 
	 }} \label{fig:afhq}
	\vspace{-4mm}
\end{figure*}
\begin{figure}[ht]
	\centering
	\includegraphics[width=0.32\textwidth]{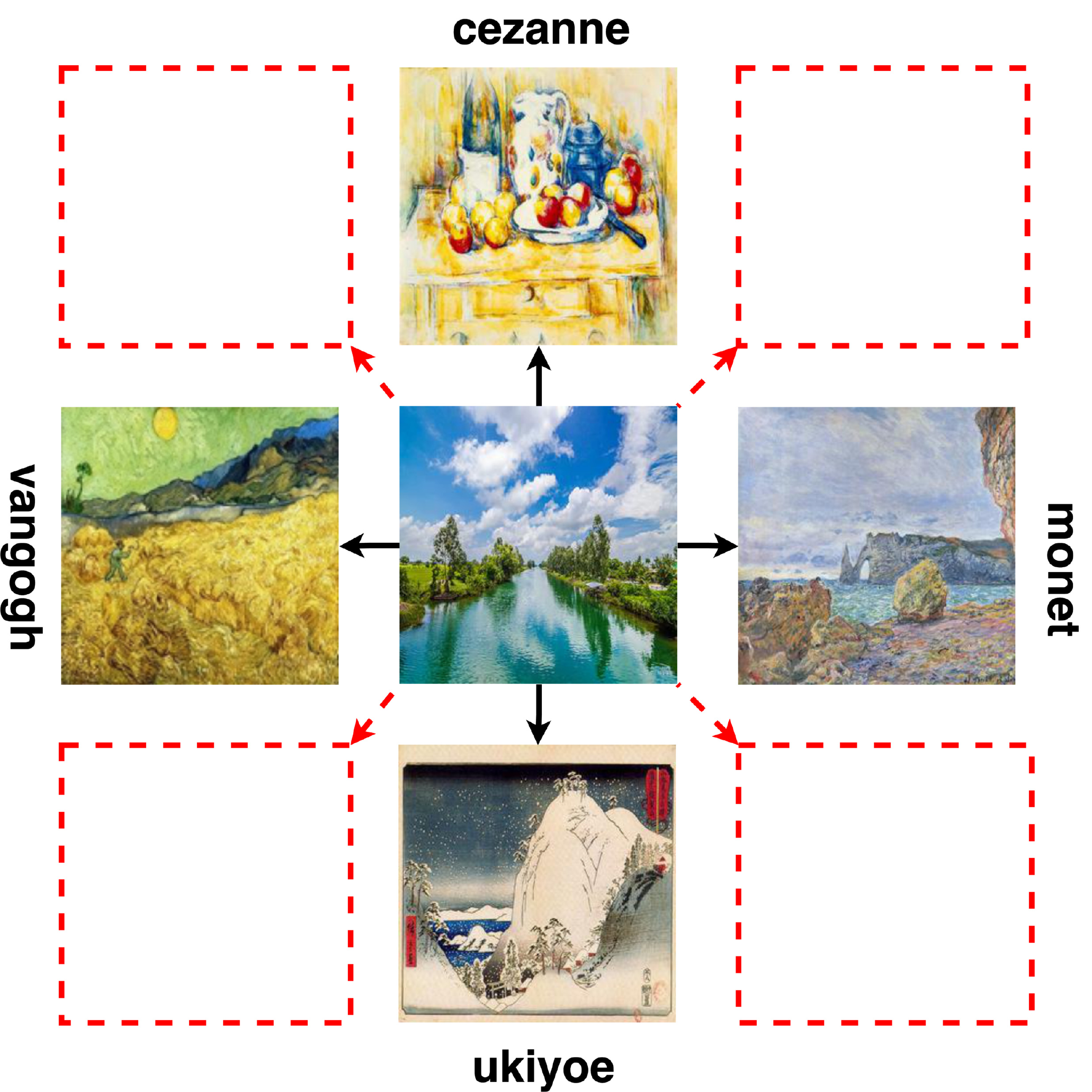}
	\vspace{-2mm}
	\caption{Example illustration of multi-domain style transfer and style generalization on the Photo2Art dataset. 
	} \label{fig:styletranfer_fusion_example}
	\vspace{-4mm}
\end{figure}
\begin{figure*}[t]
	\centering
 	\includegraphics[width=0.98\textwidth]{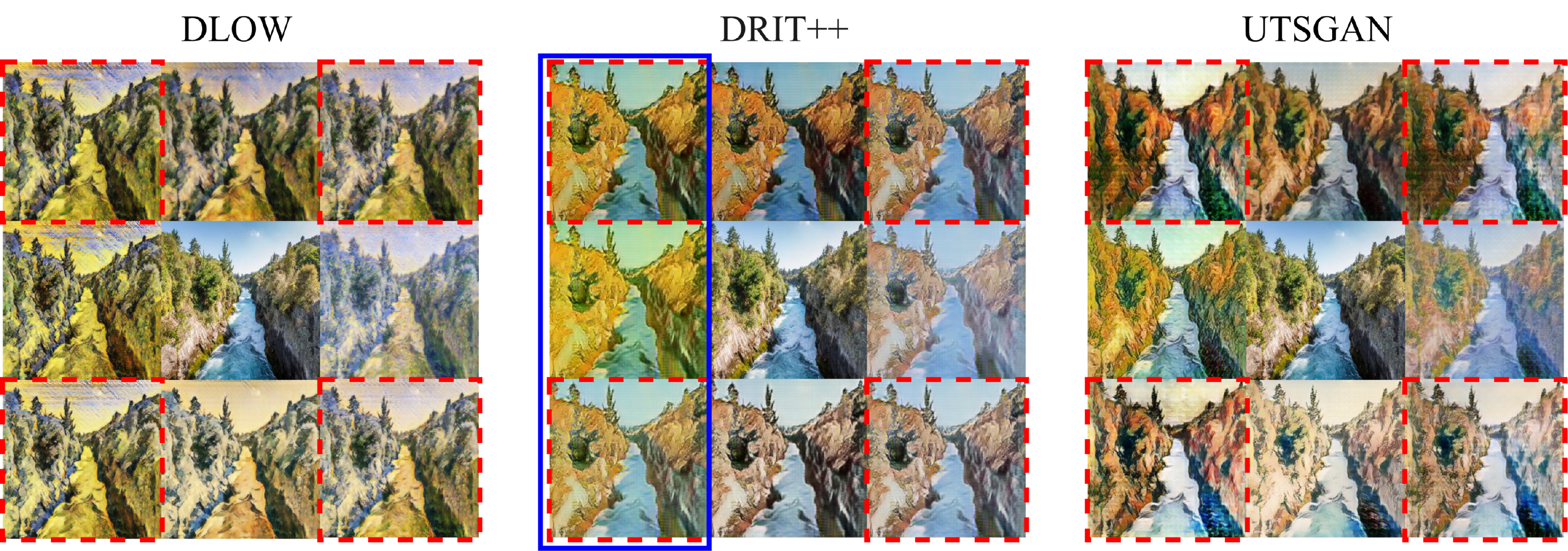}
	\vspace{-2mm}
	\caption{Comparison of DLOW, {DRIT ++} and UTSGAN on multi-domain style transfer and style generalization.  Please zoom in for details. 
	} \label{fig:styletranfer_fusion}
	\vspace{-4mm}
\end{figure*} 
%

\vspace{0.1mm}
\noindent \textbf{{Compare UTSGAN with StarGANv2:}} 
{StarGANv2~\cite{choi2020stargan} studies face editing under a different setting with UTSGAN and other baseline methods. 
Our UTSGAN aims to edit a human face so that it presents the explicitly provided semantic attributes, while StarGANv2, emphasizing diverse translation, is trained to change the input face to different general styles implicitly exemplified with different reference images.} 
For a fair comparison, we reproduce StarGANv2 under the single output domain setting and report example results in Fig.~\ref{fig:celeba_starv2}. 
The comparisons clearly show that our problem setting is different.
Take the first set as an example, while our UTSGAN subtly edits the human face to be ``\textit{Smiling}'', StarGANv2 modifies the general style of the image, with its background and skin color both look like the reference image. Such contrast is highly distinguishable in the second set, where the results of StarGANv2 present general style as its reference image, while our UTSGAN, editing images according to the specified attributes, generates results with subtle changes of ``\textit{Female}'', more ``\textit{Smiling}'' and having less ``{Mustache}''. 

\subsubsection{Multi-domain style transfer and generalization}~\label{sec:exp_style_transfer} 
We test UTSGAN on multi-domain translation tasks with two problem settings:\begin{enumerate*}
	\item[{\textit{(\romannumeral1).}}] multi-domain style transfer, \textit{i.e.}, style transfer to every single domain;
	\item[{\textit{(\romannumeral2).}}] multi-domain style generalization, \textit{i.e.}, translation with a fusion of different domains/styles. 
\end{enumerate*}

\vspace{1mm}
\noindent {\textbf{{\textit{(\romannumeral1).}} Style transfer}}

\noindent{\textbf{Evaluation on Photo2Art dataset: }} 
{In Fig.~\ref{fig:styletranfer_editing}, we compare multi-domain style transfer results on the Photo2Art dataset. 
UTSGAN produces clear and distinguishable stylization for each target domain with clear content details.
On the other hand, DRIT++ can only capture the overall style but lacks image details and shows general blurriness, as exemplified by the marked region, where the profile of bushes in the input image is mostly blurred out with the colorization.
MUNIT produces different color blends for each domain but loses significant image details, sometimes resulting in unrecognizable images~\textit{e.g.}, in the first set. 
DLOW shows clear content but fails to distinguish between target styles. 
In the second case, its results show an apparent unexpected fusion of different styles, with all four stylization results presenting evident purple color,~\textit{i.e.}, the representative color of the ``\textit{monet}'' style. 
This is because DLOW uses a linear mix-up approach for model generalization, which can disorder the model and prevent it from generating images with a clear target property.}

\noindent{{\textbf{Evaluation on AFHQ dataset: }}} 
In Fig.~\ref{fig:afhq}, the results of UTSGAN are impressive, as they present well-recognized animal faces with clear content details and distinguishable patterns for each target domain. 
The other three methods, MUNIT, DLOW, and DRIT++, all show defects in their results.
MUNIT generates results with spurious artifacts and fails to distinguish different translated domains. 
DLOW achieves reasonable results for the ``\textit{dog}'' domain but fails when stylizing cat faces to the ``\textit{wildlife}'' domain. 
DRIT++ cannot perform subtle editing on the image content for stylization, and its translated results for each domain simply show a combination of the same basic face profile with the overall color blend of the reference image, making it hard to tell the domain/breed of its generated animal faces.

In contrast, UTSGAN is able to enforce consistent translation with respect to the domain index, while still allowing in-domain variance in the output. 
This allows for animal faces of different ``breeds'' to be generated for each domain, such as tiger, fox, and others in the case of the stylization to the ``wildlife'' domain. More example results are presented in the appendix. Overall, the results demonstrate the effectiveness of UTSGAN in multi-domain style transfer.

\begin{figure*}[t]
	\centering
\includegraphics[width=0.96\textwidth]{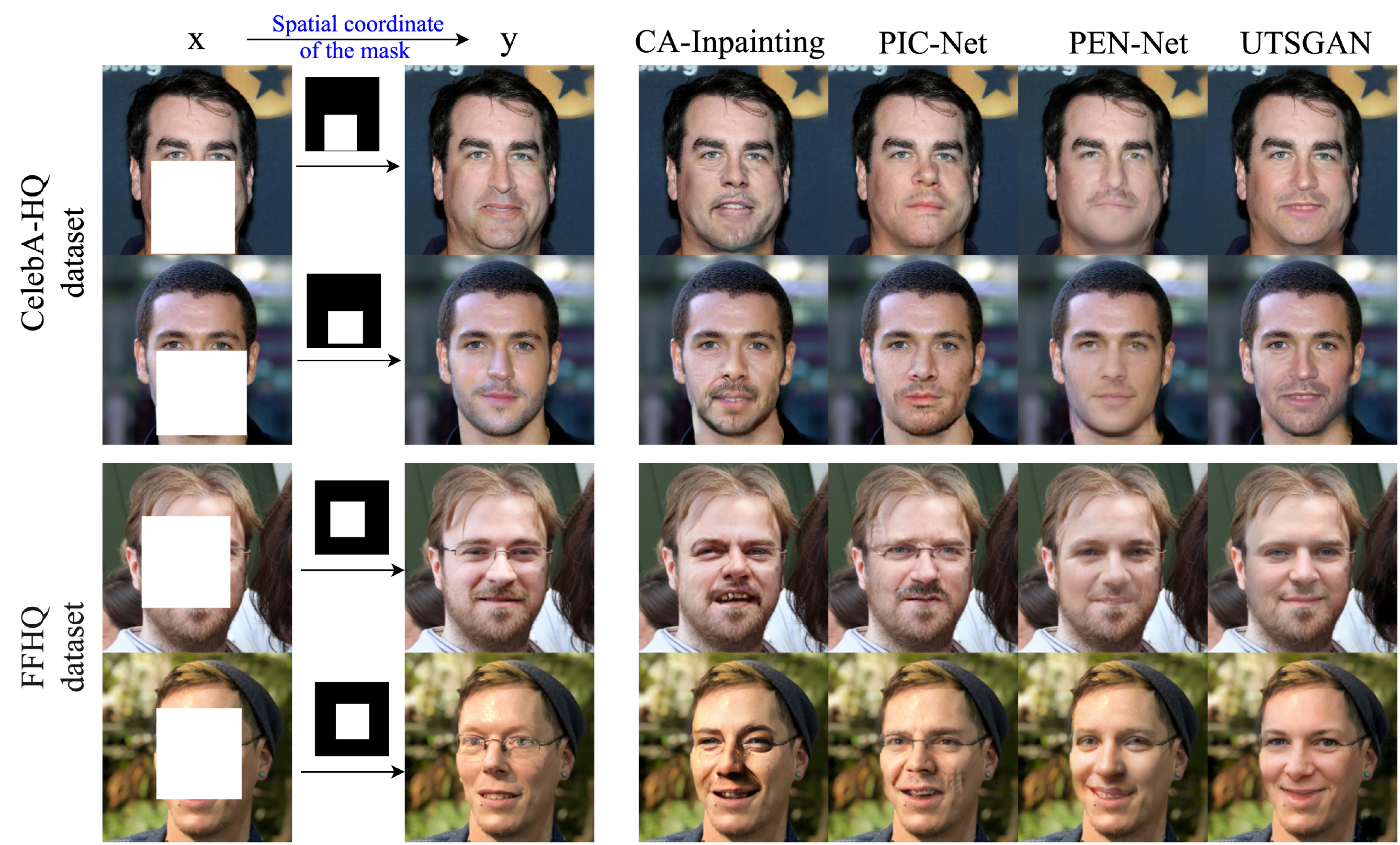}
	  \vspace{-2mm}
	\caption{Example results of CA-Inpainting, PIC-NET, PEN-NET, and our UTSGAN for image inpainting on the CelebA-HQ and the FFHQ dataset. 
} \label{fig:inpainting}
 \vspace{-4mm}
\end{figure*}

\noindent {\textbf{{\textit{(\romannumeral2).}} Style generalization}}
We further compare DLOW, DRIT++, and UTSGAN in terms of multi-domain style generalization with the Photo2Art dataset. 
Specifically, we aim to generate art paintings that incorporate a fusion of the existing observable styles. 
The example results are shown in Fig.~\ref{fig:styletranfer_fusion}, which clearly demonstrate that our UTSGAN can generate paintings that preserve clear image content with distinguishable different styles. 
More importantly, our results exhibit evident, human-understandable and smooth variance, along with the fusion of different styles.

In comparison, DLOW fails to present recognizable fused style of these target domains, although its results show some variance along with the style fusions. 
The reason is that DLOW utilizes a sample-level mix-up~\cite{DBLP:conf/iclr/ZhangCDL18} to achieve model generalization with different styles. 
However, this simplistic mix-up strategy fails to provide sufficient generalization for the model to produce smooth and easily recognizable changes across intermediate domains~\cite{DBLP:conf/icml/VermaLBNMLB19}.
The results of DRIT++ reveal a smooth variation over the fusion of different styles. 
However, its generated results fail to preserve subtle content details of the input image due to its disentanglement of content and style code. 
Moreover, its fused results do not exhibit many distinguishing style properties compared to our UTSGAN. 
For example, the pictures marked with blue box show that, apart from the color blending, DRIT++'s results of ``\textit{cazanne+ vangogh}'' and ``\textit{ukioe+ vangogh}'' do not show many differences. 
The superiority of our UTSGAN is attributed to its advantages of smooth transition mapping and generalized transition consistency. 
Specifically, it smoothly bridges the gap between different domains by learning a manifold of transition mappings to each domain. 
With generalized transition consistency on these transitions, we can thereby achieve reasonable translated results when translated with fused domain styles. 
We present more example results in the appendix.

\vspace{-2mm} 
\subsubsection{Image inpainting}\label{exp:image_inpainting}

In this subsection, we apply UTSGAN to image inpainting tasks, where the transition conveys clues of the masking regions, in each masked input (discussed in Sec.~\ref{sec:setting_inpainting}). 

Fig.~\ref{fig:inpainting} presents example results of UTSGAN and the baseline methods {on the CelebAHQ and FFHQ datasets.} \footnote{All results are direct outputs of each model without any post-processing step.} 
{For both datasets, the results of UTSGAN are the most satisfactory, as its inpainted regions are of higher quality and present a harmonious integration with the given context images. } 
The results of other methods can be analyzed as follows.

Firstly, the results of CA-inpainting are unsatisfactory for presenting observable defects with sharp and visible contrast along the boundary regions of each output image.

Secondly, the in-painted images of PIC-NET have undesired deformation of the facial organs for the given context of the image, \textit{i.e.}, the regions of the real images. 
This is because PIC-NET pursues diversity of the generation, while the consistency between the generation and the masked regions is ignored. 
In this way, although the generated images may look clear and be classified as ``real'' by the discriminator benefiting from its adopted short+long term attention layers, these images are not realistic based on human understanding.

Thirdly, the inpainting results of PEN-NET look both visually and semantically plausible compared with the other two methods since it considers these two principles simultaneously. 
However, its generated results are relatively blurred compared with the results of our UTSGAN. 
For example, 
its inpainted regions for these human faces fail to present clear details of the facial organs.

We provide more example results in the appendix. 

    
    
\begin{figure}[t]
\centering
	\includegraphics[width=0.48\textwidth]{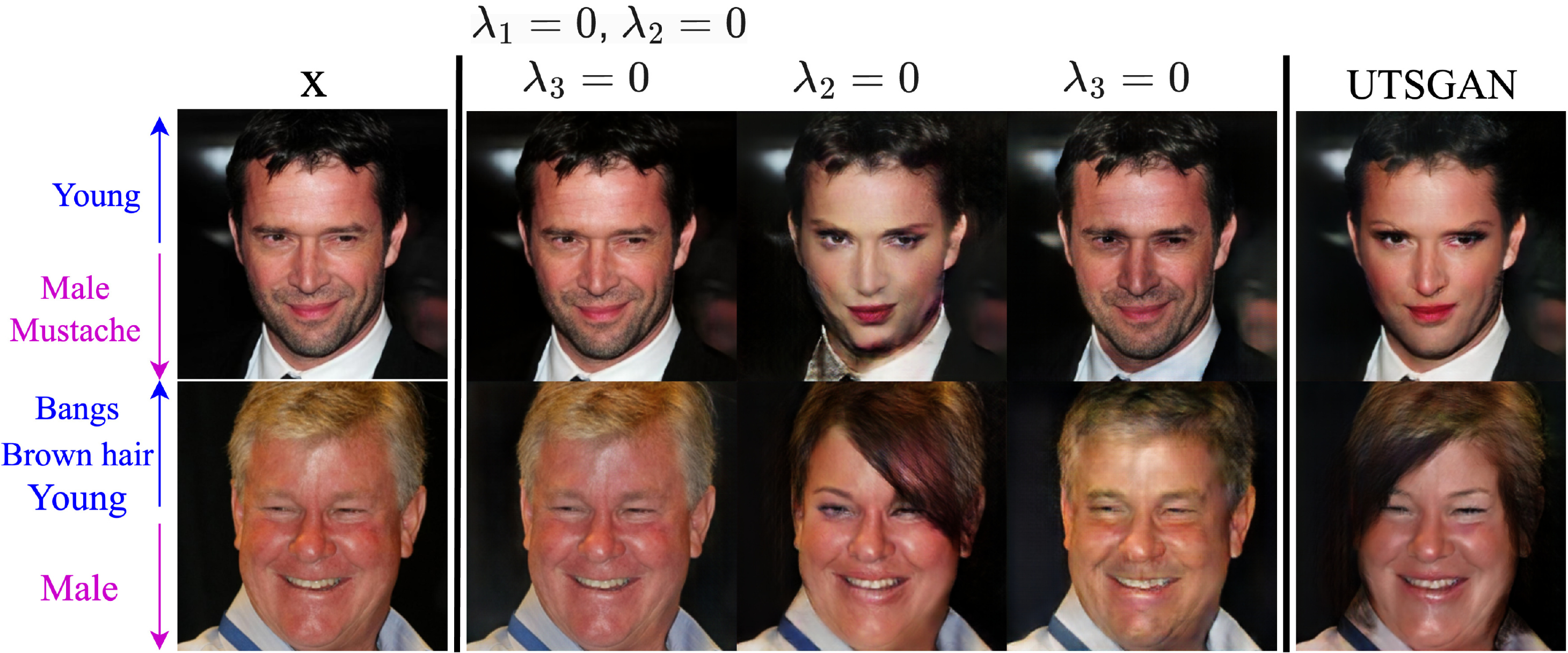} 
 	\vspace{-2mm}
  \caption{Setting and example results of ablation study. }\label{fig:ablation}
  \vspace{-5mm}
\end{figure}

\section{Ablation study}

 Based on our full objective in
 Eq.~(\ref{eq:supervised_UTSGAN}), we present an ablation study of UTSGAN with the face editing task ($\gamma=0$) under three different settings:
\begin{enumerate*}
    \item[{\textit{(\romannumeral1).}}] {\small$\lambda_1=0, \lambda_{2}=0,\lambda_{3}=0$}: the variant of UTSGAN without any regularization on transition consistency.
    \item[{\textit{(\romannumeral2).}}] {\small$\lambda_{2}=0$}: the UTSGAN variant without regularizing transition consistency on unseen transitions. 
    \item[{\textit{(\romannumeral3).}}] {\small$\lambda_3=0$}: the UTSGAN variant without generalizing the overall triplet consistency to distribution-level.
\end{enumerate*} 

{We present example results of multiple attribute editing under these three settings in Fig.~\ref{fig:ablation}. The classification accuracy corresponding to each of them is shown in Table~\ref{tab:ablation_attrs_acc}}.
It is clear that, without regularizing transition consistency (ablation1), UTSGAN mainly focuses on reconstructing the input images under the regularization of self-reconstruction and cycle-reconstruction. Its generated results fail to present the desired attribute properties, and its classification results are not satisfactory. 

Under ablation2, where the transition consistency on observed transitions is regularized and then generalized to distribution-level, the generation results successfully present the desired properties in both qualitative and quantitative aspects. However, the results under this setting are inferior to the complete UTSGAN framework. For example, in the first set, the man's ``\textit{Mustache}'' is not fully erased when translating it to a ``\textit{Female}''. The superiority of our UTSGAN indicates that the explicit regularization of transition consistency on sampled unseen transitions can benefit the model with better generalization ability for transition consistent generation. 

Under the third setting, where the joint distribution matching loss term is dropped, although the output image shows some attribute change \textit{w.r.t.} its input, the change is not what we expected. 
The comparison between ablation2 indicates that the joint-distribution matching design is essential to guarantee the superior generalization ability of UTSGAN. 

Overall, the results demonstrate that all the components in our proposed framework are essential for achieving high-quality and consistent translation results.

\begin{small}
\begin{table}[t]
\begin{center}
			\renewcommand{\arraystretch}{1}
			\caption{Classification accuracy ($\uparrow$) for the ablation of UTSGAN.}  
						\label{tab:ablation_attrs_acc}
						\vspace{-3mm}
			\setlength{\tabcolsep}{1.2mm}{
				\scalebox{1.1}{
\begin{tabular}{|c|ccc|c|}
\hline
\multirow{2}{*}{\textbf{Attributes}}& \multicolumn{4}{c|}{{\textit{\textbf{Multiple attribute editing}}}} \\
 & \multicolumn{1}{c}{\textbf{ablation 1}}&{\textbf{ablation 2}}&{\textbf{ablation 3}}&{\textbf{UTSGAN}}\\ \hline
 \textit{Bangs}      &  72.45   & 92.65             & 72.05  &  \textbf{95.70}\\
 \textit{Black\_hair}&  66.95   & \textbf{89.30}    & 68.75  &  88.60         \\
 \textit{Male}       &  55.70   & \textbf{96.75}    & 55.85  &  95.60 \\
 \textit{Mustache}   &  88.60   & 94.95             & 87.90  &  \textbf{97.25}\\
 \textit{Smiling}    &  52.85   & 91.45             & 52.30  &  \textbf{91.80} \\ \hline
 \textit{\textbf{Average}}& 67.31 & 93.02           & 67.37  & \textbf{93.79} \\ \hline
\end{tabular}
    }
}
\end{center}
\vspace{-6mm}
\end{table}
\end{small}


\section{Conclusion}
This work introduces a generative transition mechanism to enhance the traditional study of consistency in Image-to-Image (I2I) translation tasks by incorporating transition consistency. 
By combining transition consistency with result consistency and generalizing both to unseen transitions, UTSGAN surpasses previous GAN-based works in various I2I translation tasks. 
Additionally, our UTSGAN explicitly models the data mapping procedure, making it a general generative framework that can provide an explanation for existing GAN-based I2I translation models. In conclusion, our idea of transition consistency is not only insightful but also beneficial for exploring input-output consistency in other research problems.


\bibliographystyle{IEEEtran}
\bibliography{references}

\vspace{-0.5in}

\end{document}